\documentclass{ieeetmlcn}
\usepackage[dvipsnames]{xcolor}
\usepackage{subcaption}
\usepackage{adjustbox} 
\usepackage{anyfontsize}
\usepackage{makecell}
\usepackage{setspace}
\IEEEoverridecommandlockouts
\usepackage[numbers]{natbib}
\usepackage{amsmath,amssymb,amsfonts}
\usepackage{algorithmic}
\usepackage{graphicx}
\usepackage{textcomp}
\usepackage[table]{xcolor}
\usepackage{amsthm}
\usepackage{hyperref}
\usepackage{pifont}
\usepackage{marvosym}
\usepackage[most]{tcolorbox}
\usepackage{booktabs,tabulary,multirow,overpic,xcolor}
\usepackage{algorithmic}
\usepackage{algorithm}
\definecolor{darkblue}{rgb}{0.53, 0.6, 0.92} 
\definecolor{darkgreen}{rgb}{0,0.5,0}

\AtBeginDocument{\definecolor{tmlcncolor}{cmyk}{1,1,1,1}\definecolor{NavyBlue}{RGB}{0,0,0}}

\newtheorem{theorem}{Theorem}
\newtheorem{proposition}{Proposition}
\newtheorem{Proof}{Proof}

\hypersetup{
    colorlinks=true,
    linkcolor=blue,  
    urlcolor=violet,  
    citecolor=blue,
}

\newcommand{\hypbox}[1]{%
\begin{tcolorbox}[
    colback=white!98!black,
    colframe=white!30!black,
    boxsep=1.1pt,
    top=4.75pt
]%
\vspace{0.25pt}%
#1
\end{tcolorbox}
}

\usepackage{titlesec}
\titleformat{\section}{\color{black}\normalfont\large\bfseries}{\thesection}{1em}{}

\def\authorrefmark#1{\ensuremath{^{\textbf{#1}}}}

\begin{document}
\doiinfo{}

\markboth{}{}

\title{\fontsize{18.5pt}{24pt}\selectfont  \textsc{PlatoNT}: Learning a Platonic Representation for Unified Network Tomography} 

\author{Chengze Du\authorrefmark{1}, Heng Xu\authorrefmark{1}, Zhiwei Yu\authorrefmark{2}, Bo Liu\authorrefmark{1} and Jialong Li\authorrefmark{1}}
\affil{Computer Science and Control Engineering, Shenzhen University of Advanced Technology, Shenzhen, China}
\affil{Institute for Network Sciences and Cyberspace, Tsinghua University, Beijing, China}
\corresp{Corresponding author: Jialong Li (email: lijialong@suat-sz.edu.cn).}

\begin{abstract}
Network tomography aims to infer hidden network states, such as link performance, traffic load, and topology, from external observations. Most existing methods solve these problems separately and depend on limited task-specific signals, which limits generalization and interpretability. We present \textsc{PlatoNT}, a unified framework that models different network indicators (e.g., delay, loss, bandwidth) as projections of a shared latent network state. Guided by the Platonic Representation Hypothesis, \textsc{PlatoNT} learns this latent state through multimodal alignment and contrastive learning. By training multiple tomography tasks within a shared latent space, it builds compact and structured representations that improve cross-task generalization. Experiments on synthetic and real-world datasets show that \textsc{PlatoNT} consistently outperforms existing methods in link estimation, topology inference, and traffic prediction, achieving higher accuracy and stronger robustness under varying network conditions.
\end{abstract}


\maketitle

\section{INTRODUCTION}
\IEEEPARstart{N}{etwork} Tomography~\cite{vardi1996network_eng} aims to infer internal network characteristics from external observable measurements. Because direct monitoring of every internal link or node is often infeasible due to scalability, privacy, or administrative constraints, Network tomography provides a statistical and learning-based framework to reconstruct the hidden network state using limited external observations. Over the past decades, Network tomography has become a fundamental tool for understanding large-scale networks, supporting applications such as performance monitoring, anomaly detection, and network optimization~\cite{dhamdhere2007netdiagnoser, chen2003network, chen2010network}.

Classical Network tomography research has focused on three subproblems. (1) Link-level parameter inference methods~\cite{du2024identificationpathcongestionstatus, Xi2006EstimatingNL, Coates2000NetworkLI, Nguyen2007NetworkLI, zhou2024correlation} seeks to estimate unobservable link metrics, such as delay, loss rate, or available bandwidth—based on aggregated path-level measurements. (2) Origin-Destination (OD) traffic estimation methods~\cite{kumar2020multi, jiang2009garch, Polverini2018RoutingPF_od_3, Singhal2007IdentifiabilityOF_od_1} aims to infer the volume and distribution of traffic flows between source-destination pairs, providing essential inputs for capacity planning and congestion management. (3) Network topology inference methods~\cite{mle2002, EM2001-passive, ni2009efficient, gnn-nt} attempts to reconstruct the underlying structure of network nodes and links using end-to-end probing data. Together, these tasks form the core of NT, enabling indirect yet comprehensive understanding of network performance and structure.

Despite substantial progress, most existing methods rely on \textit{\textbf{limited or task-specific}} observable indicators~\cite{Nguyen2007NetworkLI, Duffield2006NetworkLT, ni2009efficient, Cceres1999MulticastbasedIO, Tsang2003NetworkDT}. A large body of work treats each NT subproblem as an independent prediction task—designing models specialized for a single metric. For instance, link loss inference models are typically trained only on path-level loss data, while delay estimation or OD traffic models are optimized on their respective measurement spaces. Such approaches overlook the inherent correlation among different indicators that collectively reflect the underlying network state, leaving their representations fragmented and weakly generalizable.

Some efforts~\cite{Papagiannaki2004ADA, zhao2006robust, kumar2020multi} attempt to mitigate these limitations by combining multiple measurements or algorithms. Multi-view fusion methods, for instance, integrate the outputs of different algorithms rather than modeling indicators jointly at the source. However, because they rely on the stability of each individual algorithm, they do not resolve the absence of a shared representation and thus generalize poorly when network conditions shift or new measurement types appear. Related studies~\cite{Bassi2022ImprovingDN_e1, Levin1998DYNAMICFE_e2, Duranthon2024AsymptoticGE_e3, du2025vaes} further show that ignoring structural dependencies in heterogeneous indicators—or in latent representations more broadly—can even harm performance. These observations highlight that simple multi-view aggregation is insufficient and reinforce the need for a unified perspective. These observations highlight that simple multi-view aggregation is insufficient and reinforce the need for a unified perspective.

We argue that the internal state of a network cannot be fully characterized by any single observable indicator. Link delay, loss, and bandwidth are not isolated properties but different manifestations of a shared latent network condition, shaped by congestion, routing, and protocol dynamics. Ignoring these interdependencies restricts each model to a limited view of the network, reducing its ability to capture the underlying physical and statistical realities. We term this limitation the \textit{\underline{\textbf{Representation Dilemma}}}: \textit{The inability of existing NT models to leverage non-target indicators to improve the accuracy and robustness of target predictions.}


To address this representation dilemma, we introduce the Plato Representation Hypothesis (PRH)~\cite{PRH_2024} as a unified framework for NT. The hypothesis is inspired by recent progress in multimodal representation learning, where models trained across diverse modalities, such as vision, language, and audio, which tend to converge toward a shared latent representation of the underlying reality. And we assume an analogous structure in NT:
\vspace{-0.15cm}
\hypbox{
    \textbf{Platonic Network Tomography} (\textsc{PlatoNT}): \textit{Observable indicators are multi-view projections of a shared latent state reflecting network performance.}
}
Under the \textsc{PlatoNT} assumption, the network condition is modeled as a latent variable, which encapsulates intrinsic network properties such as congestion levels, routing configurations, and bottleneck behaviors. Each observable indicator, such as delay or loss rate, is treated as a partial, noisy projection of this latent representation. Consequently, the objective of NT is reformulated as learning a unified latent representation that jointly explains and predicts multiple observable indicators. By training multiple downstream tomography tasks—link estimation, OD traffic prediction, and topology inference—on shared latent space, their optimization objectives impose complementary constraints, enhancing the compactness and structure of the representation.

To instantiate the PRH framework in network tomography, we design a contrastive learning-based training paradigm that aligns multi-view representations of the same network state. Each network measurement provides several correlated indicators (e.g., delay, loss, bandwidth), which are encoded into a shared latent space through indicator-specific encoders. We construct a contrastive objective~\cite{Hjelm2018LearningDR} that maximizes the mutual information between representations derived from the same underlying network state while minimizing it across different states. The loss function is theoretically grounded: we prove that contrastive learning can exactly represent the PMI kernel through inner products in the learned feature space, with an explicit bound on the shift constant derived from spectral theory (\textit{Theorem~\ref{thm: pmi_kernel}}). Furthermore, we show that joint optimization in the shared latent subspace yields a $1/r$ reduction in composite gradient magnitude, where $r$ is the subspace dimension (\textit{Proposition~\ref{prop: gradient_reduction}}). Together, these results establish a principled and trainable framework for unified multi-indicator modeling in network tomography.

\textbf{Contributions}. We establish a unified framework for network tomography based on the Plato Representation Hypothesis. This work \textbf{re-examines} network tomography from a multi-view representation learning perspective and provides both theoretical and practical foundations for this formulation. The main contributions of this paper are summarized as follows:

\begin{itemize}
    \item We reinterpret delay, loss, and bandwidth as different projections of a shared latent network state, providing a unified representation view that connects heterogeneous network indicators.
    \item We design a contrastive learning objective with theoretical guarantees, proving that it exactly captures the PMI kernel and ensuring principled multi-indicator alignment within a shared latent space.
    \item We develop a unified training framework combining denoising reconstruction and multi-task regularization, leading to more stable optimization and superior performance on link estimation, OD, and topology inference.
\end{itemize}



 \begin{figure}[t]
	\centering
	\includegraphics[width=0.95\linewidth]{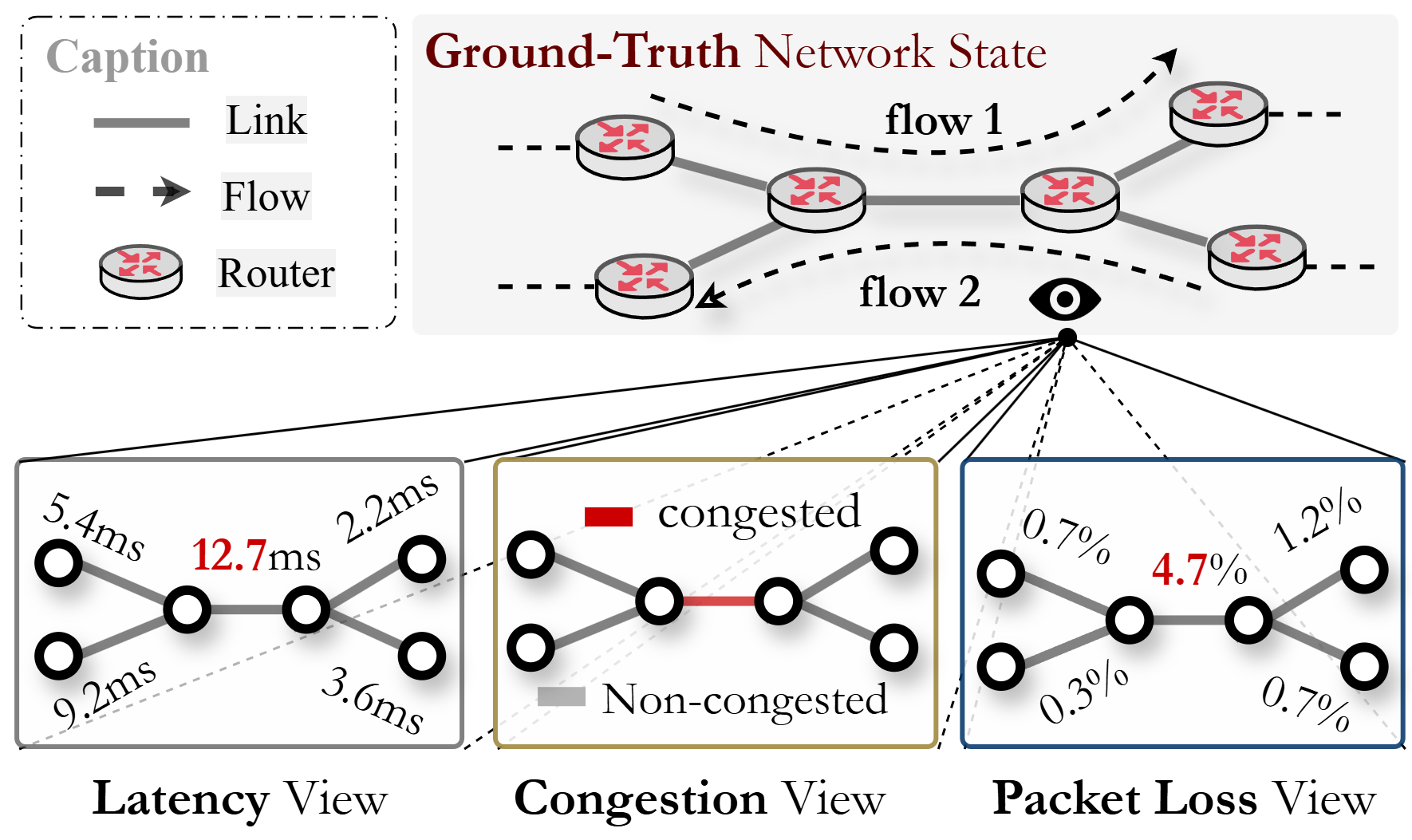}
     \caption{Illustration of \textbf{Platonic Network tomography}. Latency, congestion, and loss views represent distinct yet complementary projections of a shared latent network state.}
     \vspace{-0.5cm}
	\label{intro_PRH}
\end{figure}

\textbf{Organization.} The remainder of this paper is organized as follows. We first review related work (Section~\ref{sec: related_works}) and theoretical background (Section~\ref{sec: bck}), then introduce our method (Section~\ref{sec: method}), design (Section~\ref{sec: design}) and experimental results (Section~\ref{sec: exp}), followed by concluding remarks (Section~\ref{sec: conclusion}). Proofs are included in the Appendix~\ref{sec: appendix1} for completeness.

\section{Related Works}
\label{sec: related_works}

We review prior work on three core network tomography subproblems—link-level performance estimation, OD matrix inference, and topology inference—and conclude with recent progress in contrastive learning relevant to our approach.

\textit{Link-level Parameter Estimation} in network tomography focuses on deducing link states or performance metrics from end-to-end path measurements. Early analog methods, such as those in \cite{ant-1-Caceres_Duffield_Moon_Towsley_2003, chen2010network, ant-5-Ghita_Nguyen_Kurant_Argyraki_Thiran_2010,ant-3-Sossalla,ant-4-qiao,add-new-kolar2020distributed, Coates2000NetworkLI, Xi2006EstimatingNL}, use invasive probing to solve under-constrained linear equations, often incorporating optimization constraints to improve accuracy, though they demand significant computational and probing resources \cite{NBT_Duffield_2006}. Boolean tomography, explored in \cite{nbt-1-Duffield_2003,nbt-2-ogino,nbt-4-bartolini,NBT_Duffield_2006}, employs binary path measurements to infer link status (e.g., 'Bad' or 'Normal'). Range tomography \cite{range_nt} extends this by identifying both link status and congestion ranges to accommodate diverse network service requirements. Recent advances leverage machine learning, with works like \cite{add-new-nt-dl-hudeepnt,add-new-nt-dl-sartzetakis2022machine,add-new-nt-dl-tao2023network, ma2020neuralnetworktomography} using Neural Networks to enhance scalability and adaptability in inferring network structures or predicting path performance, while some studies \cite{du2025roto, secureNT} address vulnerabilities to attacks that manipulate inferred results by increasing link delays.

\textit{OD matrix inference}, or traffic matrix estimation (TME), and topology inference address complementary subproblems in network tomography. TME estimates source-destination traffic flows from indirect measurements like link loads, tackling ill-posed inverse problems where OD pair counts exceed link counts, requiring additional assumptions or data \cite{chen2017efficient, xu2019lightweight, zhou2018highly}. Early approaches \cite{zhang2003fast, erramill2006independent} assume constant or independent traffic ratios, while sparse methods \cite{mardani2015estimating, soule2005traffic} exploit low-rank or sparse TM properties via compressed sensing to constrain solutions. Statistical models \cite{jiang2009garch, xu2021learning} incorporate traffic distribution assumptions, and others \cite{liu2018tomogravity} address measurement noise to improve accuracy, with extensions handling heterogeneous measurements \cite{zhao2006robust} and software-defined networks \cite{tian2018sdn} to reduce overhead and bias. \textit{Topology inference} reconstructs internal network structures from end-to-end delay measurements without accessing intermediate nodes. Early methods like MLE \cite{mle2002} and EM \cite{EM2001-passive} use additive delay models to identify bottlenecks, while algebraic approaches \cite{ni2009efficient} handle sparse data, and recent deep learning techniques \cite{gnn-nt} map delays to topologies directly, improving robustness and scalability. 

\textit{Contrastive Learning} (CL)~\cite{chen2020simpleframeworkcontrastivelearning, Hjelm2018LearningDR, wu2018unsupervisedfeaturelearningnonparametric}, a self-supervised method, approximates data generation reversal by aligning related samples and separating unrelated ones in representation space, recovering identifiable latent factors without strong independence assumptions \cite{c-12,c-13}. Contrastive-equivariant frameworks enhance alignment by preserving input transformation variability, applied to network event correlation and temporal dependency capture \cite{c-14,c-15,c-16, du2025semitoneaware}. The Platonic Representation Hypothesis (PRH)~\cite{PRH_2024} suggests deep neural networks across architectures, objectives, and modalities converge to a shared statistical model of underlying reality, with observed data as its projections. The Multi-Task Scaling Hypothesis (MTSH) states that increasing task complexity constrains representation space, forcing convergence to general representations, guiding shared low-level feature spaces in multi-task learning, as in unified architectures like PixelBytes \cite{c-21,c-23,c-27}.
Some methods~\cite{soule2005traffic, zhao2006robust, Malboubi2018OptimalCoherentNI} address network issues from a multi-source perspective, but their approaches often rely on simple feature concatenation (like vertical concatenation in RTME~\cite{zhao2006robust}), which can lose the statistical properties of each data source. Other multi-view approaches~\cite{kumar2020multi} use CCA, but they cannot distinguish individual signals and force orthogonal subspaces, which does not align with observed resource data. Several works~\cite{11119637,10.1145/3651890.3672245} establish unified frameworks for optical networks, but they target control-plane abstraction or physical-layer optimization, which are fundamentally different from the network tomography problem.

\section{Backgrounds}
\label{sec: bck}
We outline the background of our framework, covering the measurement–measurand formulation of NT and the mutual-information perspective of contrastive learning.
\vspace{-0.45cm}
\subsection{Network Model and Tomography}
Network tomography (NT) provides a mathematical framework for inferring unobservable internal network quantities from externally measurable data. 
Let $\mathcal{G}=(\mathcal{V},\mathcal{L})$ denote a network consisting of node set $\mathcal{V}$ and link set $\mathcal{L}$. 
Denote by $\mathbf{x}\in\mathbb{R}^{|\mathcal{L}|}$ the vector of \emph{measurands}—the hidden internal quantities of interest—and by $\mathbf{y}\in\mathbb{R}^{|\mathcal{P}|}$ the vector of \emph{measurements} collected along a set of probing paths $\mathcal{P}$.

The mapping between $\mathbf{x}$ and $\mathbf{y}$ is governed by the \emph{routing matrix} $\mathbf{R}\in\{0,1\}^{|\mathcal{P}|\times|\mathcal{L}|}$, where $R_{ij}=1$ indicates that link $j$ lies on probing path $i$, and $R_{ij}=0$ otherwise. 
Given $\mathbf{R}$, the measurement model can be expressed as
\begin{equation}
\begin{aligned}
    \mathbf{Y} = \mathbf{A}(\mathbf{R}, \mathbf{X}) + \boldsymbol{\epsilon} 
     \ \Rightarrow  \ 
    y_{i} = \odot_{j=1}^{|\mathcal{L}|} R_{ij} \cdot x_{j} + \epsilon_{i},
\end{aligned}
\end{equation}
where $\boldsymbol{\epsilon}$ denotes measurement noise, and the operator $\odot$ represents a generalized path-wise aggregation operator. 
Specifically, $\odot$ abstracts the combination rule of link-level quantities along a path, which can be \textit{additive} (e.g., delay summation), \textit{multiplicative} (e.g., end-to-end loss rate), or even \textit{Boolean} (e.g., link failure status). 


For different NT subtasks, each one corresponding to an interpretation of the measurand $\mathbf{x}$ and measurement $\mathbf{y}$.

\vspace{0.1cm}
\noindent
\textbf{(a) Link-level Parameter Estimation.}
This task seeks to infer per-link performance parameters (e.g., delay, loss, bandwidth) from end-to-end aggregated measurements. 
Here, the measurand $\mathbf{x}$ represents link-level quantities, while $\mathbf{y}$ denotes measurable path-level data.
Under an additive model, the relationship is given by
    $
    \mathbf{y} = \mathbf{R}\mathbf{x} + \boldsymbol{\epsilon},
    $
where $\mathbf{R}$ encodes the path–link incidence relation. 
Since $\mathbf{R}$ is typically rank-deficient, this inverse problem is ill-posed, and additional regularization or learning priors are required to recover $\mathbf{x}$.

\vspace{0.1cm}
\noindent
\textbf{(b) Origin–Destination (OD) Traffic Estimation.}
OD tomography infers the traffic volumes exchanged between source–destination pairs. 
In this case, the measurand $\mathbf{x}$ corresponds to OD flows, and the measurement $\mathbf{y}$ corresponds to link load data, such that
$
    \mathbf{y} = \mathbf{R}\mathbf{x} + \boldsymbol{\epsilon}.
$
Because the number of OD pairs often exceeds the number of measured links, the system is underdetermined, requiring structural assumptions such as temporal smoothness, low-rankness, or sparsity.

\vspace{0.1cm}
\noindent
\textbf{(c) Topology Inference.}
Topology inference aims to reconstruct the network connectivity itself. 
Unlike the previous two tasks, the routing matrix $\mathbf{R}$ is unknown and must be estimated jointly with $\mathbf{x}$. 
This task is often formulated as a probabilistic optimization problem:
\begin{equation}
    T^{*} = \arg\max_{T \in \mathcal{R}} p(\mathbf{y, x}, T),
\end{equation}
where $T$ denotes a candidate topology in the feasible set $\mathcal{R}$. 
Inference methods include maximum-likelihood estimation, Bayesian inference, or deep learning–based structure recovery from end-to-end delay statistics.

In all three subtasks, the central objective is to reconstruct the hidden network quantities from indirect and noisy observations. 
Despite differing in observables and parameterization, they share the same fundamental measurement–measurand paradigm, which motivates unified learning-based formulations for network tomography.
\vspace{-0.4cm}
\subsection{Contrastive Learning}

Contrastive learning (CL) aims to learn representations by maximizing the mutual information (MI) between related samples while minimizing it across unrelated ones. 
Formally, the mutual information between two random variables $X$ and $Y$ is defined as the Kullback–Leibler (KL) divergence between their joint distribution $P_{XY}$ and the product of their marginals $P_X P_Y$:
\begin{equation}
    I(X;Y) = D_{KL}\!\left(P_{XY} \, \| \, P_X P_Y\right)
    = \mathbb{E}_{P_{XY}}\!\left[\log \frac{p(x,y)}{p(x)p(y)}\right].
\end{equation}

Directly computing $I(X;Y)$ is generally intractable because $p(x,y)$ and $p(x)p(y)$ are unknown and difficult to estimate in high-dimensional settings. 
To obtain a tractable objective, the KL divergence can be lower-bounded using the \textit{Donsker–Varadhan (DV) representation}~\cite{Donsker1975AsymptoticEO}:
\begin{equation}
    D_{KL}(P\|Q) \ge
    \sup_{T_\theta} \, 
    \Big(
    \mathbb{E}_{P}\!\left[T_\theta(x,y)\right]
    - \log \mathbb{E}_{Q}\!\left[e^{T_\theta(x,y)}\right]
    \Big).
\end{equation}

where $T_\theta(x,y)$ is a function parameterized by $\theta$ (seen as a neural network) that approximates the log-density ratio $\log \frac{p(x,y)}{p(x)p(y)}$. 
Substituting this into the mutual information definition yields the \textit{DV lower bound of mutual information}:
\begin{equation}
\begin{aligned}
    I&(X;Y) \ge I_{DV}^{\text{sup}}(X;Y)\\
    & = \sup_{\theta}\!
    \left(
    \mathbb{E}_{p(x,y)}[T_\theta(x,y)]
    - \log \mathbb{E}_{p(x)p(y)}[e^{T_\theta(x,y)}]
    \right).
\end{aligned} 
\end{equation}

Maximizing this lower bound encourages $T_\theta(x,y)$ to assign higher scores to samples drawn from the true joint distribution $p(x,y)$ (i.e., \textit{positive pairs}) and lower scores to samples drawn independently from $p(x)p(y)$ (i.e., \textit{negative pairs}). 
This formulation naturally leads to a binary classification view: distinguishing whether a pair $(x,y)$ originates from the joint or the product of marginals. 
Practically, the contrastive loss function implements this discrimination process by contrasting positive and negative pairs, for example through the InfoNCE~\cite{He2019MomentumCF} objective or its variants.

\section{Methodology}
\label{sec: method}
This section outlines the methodology of \textsc{PlatoNT}, including the problem formulation, key definitions for objective construction, and the final learning objective.

\subsection{Problem Formulation}

We begin by formalizing the multi-view network tomography problem. Consider a network $G = (V, L)$ with node set $V$ and link set $L$. For each probing path or network segment, we observe multiple correlated indicators that reflect different aspects of network performance.

Let $\mathbf{x}^{(i)} \in \mathbb{R}^{d_i}$ denote the $i$-th indicator type (e.g., delay, loss rate, bandwidth), where $d_i$ is the dimension of that indicator. In network tomography, these indicators are not independent observations but rather different manifestations of the same underlying network state. \textbf{Our goal} is to learn a unified latent representation $\mathbf{z} \in \mathbb{R}^d$ that captures the shared network condition across all observable indicators.

\subsection{Foundational Concepts}

To quantify the relationship between different network indicators, we define a kernel function that measures their similarity with respect to the underlying network state. 

Consider network measurements collected over time. Let $\mathbf{x}^{(i)}_t$ denote the measurement of indicator type $i$ at time $t$, and $\mathbf{x}^{(j)}_{t'}$ denote the measurement of indicator type $j$ at time $t'$. We define the \textbf{\textit{kernel function}} as:
\begin{equation}
    k(\mathbf{x}^{(i)}_t, \mathbf{x}^{(j)}_{t'}) = \exp\left(-\frac{\|\phi_i(\mathbf{x}^{(i)}_t) - \phi_j(\mathbf{x}^{(j)}_{t'})\|^2}{2\sigma^2}\right),
    \label{eq: kernal_func}
\end{equation}

where $\phi_i: \mathbb{R}^{d_i} \rightarrow \mathbb{R}^d$ is an encoder that maps indicator $i$ into the shared latent space, and $\sigma$ is a temperature parameter controlling the sensitivity of similarity. This kernel captures the intuition that indicators derived from the same network state at the same or nearby time should yield similar latent representations.

To formalize the relationship between indicators, we introduce a temporal dimension into the co-occurrence framework. Network indicators are not static but evolve over time as network conditions change. We define the temporal co-occurrence probability within a time window $[t, t+\Delta t]$.

Let $\mathbf{x}^{(i)}_t$ and $\mathbf{x}^{(j)}_t$ be two indicator types measured at time $t$ on the same network element (path or link). Their temporal co-occurrence is governed by the joint distribution $p(\mathbf{x}^{(i)}_t, \mathbf{x}^{(j)}_t | \mathbf{z}_t)$, where $\mathbf{z}_t$ is the latent network state at $t$.

We define the \textbf{temporal co-occurrence event} as:
\begin{equation}
    C_{ij}(t) = \{\mathbf{x}^{(i)}_t, \mathbf{x}^{(j)}_t \text{ within } [t, t+\Delta t]\}.
\end{equation}

The \textbf{temporal co-occurrence probability} is then:
\begin{equation}
    p(C_{ij}(t)) = \int_{\mathcal{Z}} p(\mathbf{x}^{(i)}_t | \mathbf{z}_t) p(\mathbf{x}^{(j)}_t | \mathbf{z}_t) p(\mathbf{z}_t) d\mathbf{z}_t,
\end{equation}

where $p(\mathbf{x}^{(i)}_t | \mathbf{z}_t)$ represents the conditional distribution of indicator $i$ given the latent state at time $t$. This formulation gives insight that indicators co-occurring on the same network element within a time window share a common latent cause, reflecting the instantaneous network condition.


To measure the statistical dependency between indicators beyond simple co-occurrence, we employ the pointwise mutual information (PMI) kernel. For two indicator samples $\mathbf{x}^{(i)}_t$ and $\mathbf{x}^{(j)}_{t'}$ measured at times $t$ and $t'$, PMI is defined as:
\begin{equation}
    \text{PMI}(\mathbf{x}^{(i)}_t, \mathbf{x}^{(j)}_{t'}) = \log \frac{p(\mathbf{x}^{(i)}_t, \mathbf{x}^{(j)}_{t'})}{p(\mathbf{x}^{(i)}_t)p(\mathbf{x}^{(j)}_{t'})}.
    \label{eq: pmi}
\end{equation}

The PMI quantifies how much more (or less) likely two indicators co-occur compared to what would be expected under statistical independence. A positive PMI indicates that the indicators are more likely to appear together than by chance, suggesting a shared underlying cause.

We extend this to define the \textbf{PMI kernel} $K_{\text{PMI}}$ over the latent representations:
\begin{equation}
    K_{\text{PMI}}(\mathbf{z}_t, \mathbf{z}_{t'}) = \mathbb{E}_{\mathbf{x}^{(i)}, \mathbf{x}^{(j)} \sim p(\cdot | \mathbf{z}_t, \mathbf{z}_{t'})} \left[\text{PMI}(\mathbf{x}^{(i)}, \mathbf{x}^{(j)})\right],
    \label{eq: pmi_kernel}
\end{equation}

which measures the expected PMI between indicators generated from latent states $\mathbf{z}_t$ and $\mathbf{z}_{t'}$.

\subsection{Unified Representation Objective}

Under the Platonic Network Tomography framework, our objective is to construct a latent representation that captures the mutual information between different views of the same network state. We formalize this through the mutual information between latent representations of different indicators.

For indicators $i$ and $j$, let $\mathbf{Z}^{(i)} = \phi_i(\mathbf{X}^{(i)})$ and $\mathbf{Z}^{(j)} = \phi_j(\mathbf{X}^{(j)})$ denote their latent representations. 
Given the temporal co-occurrence structure in network measurements, we can decompose the mutual information based on positive and negative pairs. Let $(\mathbf{x}^{(i)}_t, \mathbf{x}^{(j)}_t)$ denote a positive pair—indicators measured on the same network element at time $t$—and $(\mathbf{x}^{(i)}_t, \mathbf{x}^{(j)}_{t'})$ with $t \neq t'$ denote a negative pair from different time periods or network elements. The mutual information can be expressed as:
\begin{equation}
    I(\mathbf{Z}^{(i)}; \mathbf{Z}^{(j)}) = \mathbb{E}_{(\mathbf{z}^{(i)}_t, \mathbf{z}^{(j)}_t) \sim p_+} \left[\log \frac{p(\mathbf{z}^{(i)}_t, \mathbf{z}^{(j)}_t)}{p(\mathbf{z}^{(i)}_t)p(\mathbf{z}^{(j)}_t)}\right],
\end{equation}

where $p_+$ is the distribution of positive pairs. High mutual information indicates that knowing one representation provides substantial information about the other, which is desirable when both stem from the same underlying network state. This formulation naturally connects to the PMI kernel: the objective seeks representations where the expected PMI over co-occurring indicators is maximized.

For practical implementation, we approximate the distributions using empirical samples. Assuming that given a batch of $N$ samples at different time points, we estimate the mutual information through:
\begin{equation}
    I(\mathbf{Z}^{(i)}; \mathbf{Z}^{(j)}) \approx \frac{1}{N} \sum_{n=1}^N \log \frac{s(\mathbf{z}^{(i)}_{t_n}, \mathbf{z}^{(j)}_{t_n})}{\frac{1}{N}\sum_{m=1}^N s(\mathbf{z}^{(i)}_{t_n}, \mathbf{z}^{(j)}_{t_m})},
    \label{eq: app_mi}
\end{equation}

where $s(\mathbf{z}^{(i)}, \mathbf{z}^{(j)})$ is a similarity function and $t_n$ denotes the time of the $n$-th sample. The numerator represents the similarity between co-occurring indicators at the same time (positive pairs), while the denominator normalizes over all possible temporal pairings, effectively incorporating both positive and negative pairs.

Formulation\,\ref{eq: app_mi} exhibits several key properties that benefit network tomography. Through mutual information, it promotes alignment between representations from different indicators of the same network state, while the normalization term ensures discrimination between representations from distinct network states or time periods. Lastly, its expectation-based formulation enables efficient computation without the need for exhaustive enumeration of all pairs.

The contrastive learning objective in Eq.~(\ref{eq: app_mi}) is theoretically grounded by the following result:

\begin{theorem}[Exact PMI Kernel Representation]
\label{thm: pmi_kernel}
There exists a feature map $f_X: \mathcal{X} \to \mathbb{R}^d$ such that
\begin{equation}
\langle f_X(x_i), f_X(x_j) \rangle = K_{\text{PMI}}(x_i, x_j) + C,
\end{equation}
where $K_{\text{PMI}}$ is the PMI kernel defined in Eq.~(\ref{eq: pmi_kernel}). The shift constant $C$ admits an explicit bound:
\begin{equation}
C \geq \max\left(0,\ (N-1)\alpha - \min_i K_{ii}\right),
\end{equation}
where $\alpha = \max_{i\neq j}|K_{ij}|$ captures the maximum absolute value of off-diagonal PMI entries, and $K_{ii}$ denotes the diagonal PMI values. Under smoothness assumptions on the indicator co-occurrence distribution (detailed in Appendix~\ref{sec: appendix1}), if the co-occurrence probability satisfies $\rho_{\min} \in (0,1]$ and the off-diagonal variation parameter $\varepsilon$ satisfies $\varepsilon \geq N|\log \rho_{\min}|$, then $K$ itself is positive semi-definite.
\end{theorem}


\noindent
Theorem~\ref{thm: pmi_kernel} establishes that our contrastive objective learns representations that exactly capture the statistical dependencies between network indicators, providing theoretical justification for the alignment loss.

\section{\textsc{PlatoNT}  Design}
\label{sec: design}
This section presents the architecture and training methodology of \textsc{PlatoNT}. We begin by articulating the design intuition grounded in the PRH, then describe the model architecture with multi-indicator encoders and task-specific decoders, and finally detail the unified learning objective.
 \begin{figure}[t]
	\centering
	\includegraphics[width=\linewidth]{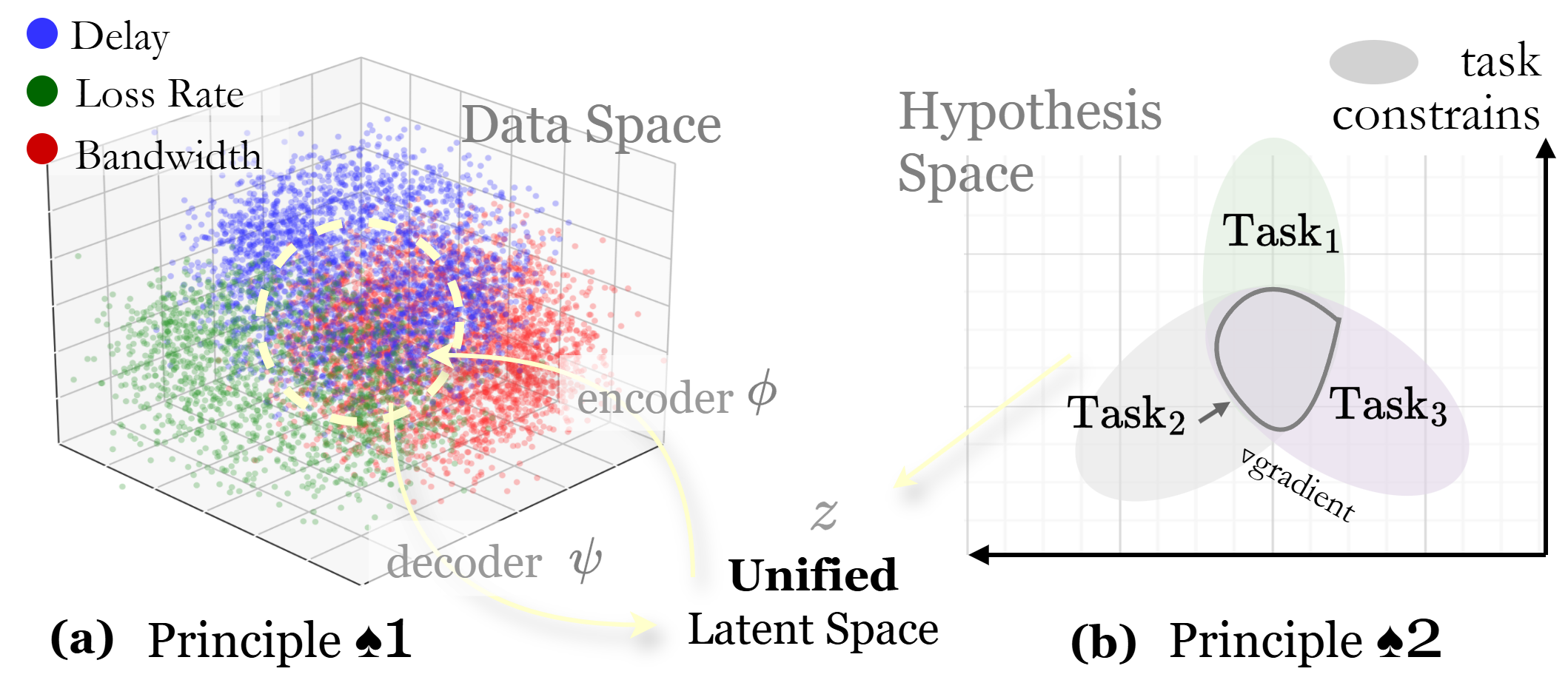}
     \vspace{-0.2cm}
     \caption{\textsc{PlatoNT} design intuition. \textbf{(a)} Multi-indicator alignment: Network indicators converge to a shared latent representation through encoder $\phi$. \textbf{(b)} Denoising \& tasks regularization: The decoder $\psi$ reconstructs clean indicators from the latent space, then used for task-specific algorithms.}
     \vspace{-0.6cm}
	\label{fig: intuition}
\end{figure}

\subsection{Design Intuition}

The design of \textsc{PlatoNT} (seen in Figure\,\ref{fig: intuition}) is guided by the PRH, which yields two complementary design principles.

\vspace{0.05cm}
$\spadesuit \mathbf{1}$ \underline{\textit{\textbf{Explicit multi-indicator alignment}}}: Different observable metrics—such as delay, loss, and bandwidth—should converge in the latent space if they originate from the same network condition. This alignment enforces the latent variable $z$ to encode consistent information across modalities, reducing redundancy and enhancing identifiability. By treating each indicator as a distinct view of the underlying network state, we construct a shared representation that captures the invariant properties governing network performance.

\vspace{0.05cm}
$\spadesuit \mathbf{2}$ \underline{\textit{\textbf{Denoising and Implicit tasks regularization}}}: The reconstruction loss with clean and noisy samples enables the model to learn denoised representations of network indicators. By reconstructing clean ground-truth from noisy measurements, the decoder learns to filter out measurement noise while preserving essential network state information. These denoised indicators can then be directly used by downstream tomography algorithms.

Fundamental \textbf{insight}: the true network condition manifests across multiple observable dimensions and multiple inference tasks. By aligning indicators in the representation space and regularizing through multi-task learning, \textsc{PlatoNT} learns a unified latent state that is both physically grounded and computationally effective.

\subsection{Architecture Overview}

\begin{figure}[t]
	\centering
	\includegraphics[width=\linewidth]{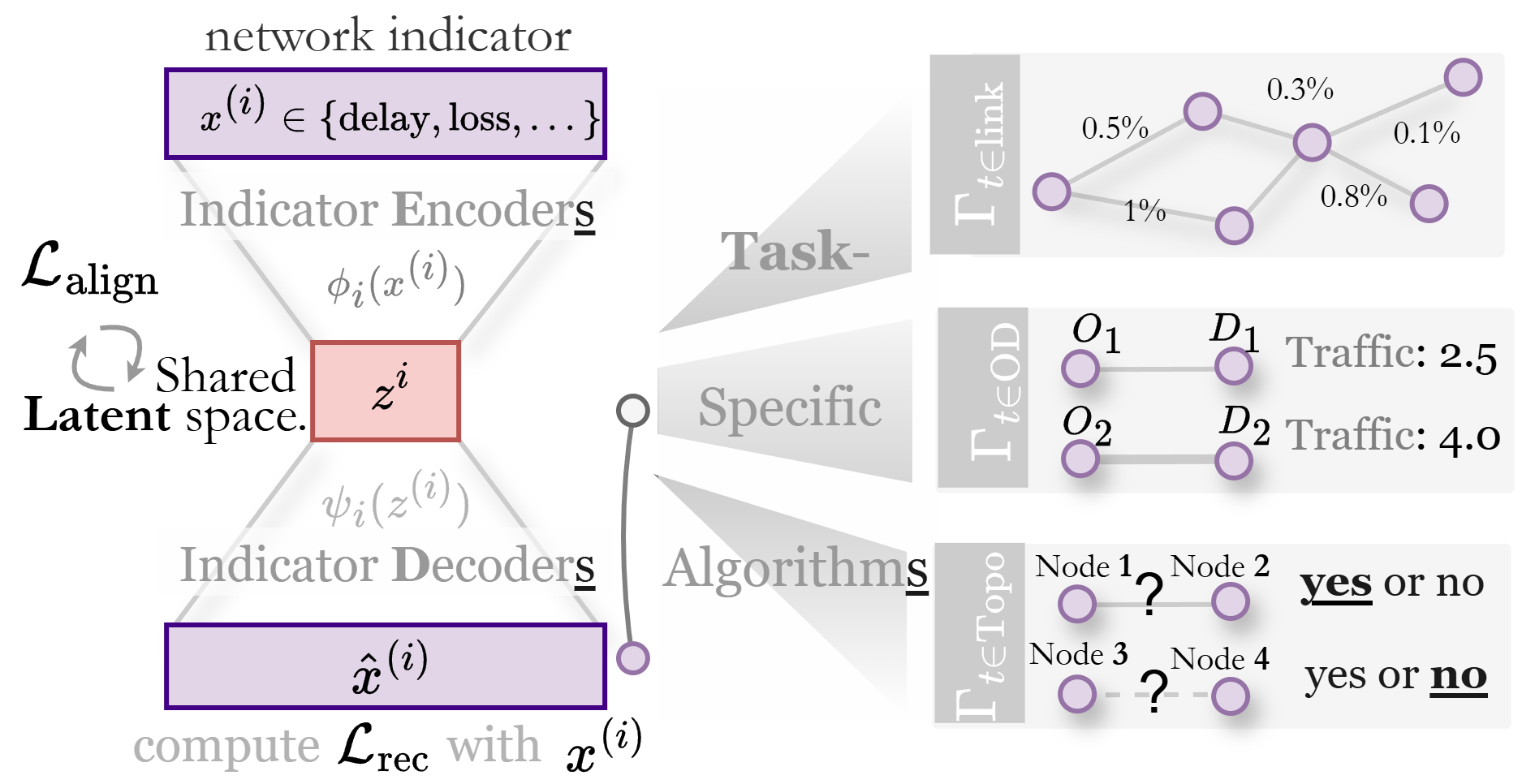}
     \vspace{-0.3cm}
     \caption{Design of \textsc{PlatoNT}. Multiple network indicators are encoded into a shared latent space, aligned via $\mathcal{L}_{\text{align}}$. 
    The latent representation is reconstructed by indicator decoders and the denoised indicators are then used by task-specific algorithms for downstream tomography tasks.} 
     \vspace{-0.8cm}
	\label{fig: design}
\end{figure}

The architecture of \textsc{PlatoNT} (seen in Figure\,\ref{fig: design}) consists of three main components: indicator encoders, a shared latent representation space, and indicator decoders. 

\textbf{Indicator Encoders.} Each network indicator $x^{(i)} \in \mathbb{R}^{d_i}$ (where $i \in \{\text{delay}, \text{loss}, \text{bandwidth}, ...\}$) is processed by a dedicated encoder $\phi_i: \mathbb{R}^{d_i} \rightarrow \mathbb{R}^d$. These encoders map raw measurements into a shared $d$-dimensional latent space:
$
z^{(i)} = \phi_i(x^{(i)}),
$
where $z^{(i)} \in \mathbb{R}^d$ represents the latent encoding of indicator $i$. Each encoder is implemented as a multi-layer perceptron with ReLU activations, allowing it to learn nonlinear transformations specific to its input modality while targeting a common representation space.

After encoding different network indicators into the shared latent representation $z \in \mathbb{R}^d$, the subsequent multi-task optimization exhibits favorable gradient properties. Recall that task-specific decoders $\Gamma_t$ map from this shared space to different prediction targets. The following result explains why joint optimization in this shared latent space improves convergence.

\textbf{Shared Latent Representation.} The latent space $\mathcal{Z} \subset \mathbb{R}^d$ serves as the Platonic representation of network state. Under the PRH, indicators derived from the same network condition at time $t$ should yield similar latent vectors:
$
z^{(i)}_t \approx z^{(j)}_t \quad \text{for all } i, j,
$
where the similarity is measured by the kernel function defined in Eq.~(\ref{eq: kernal_func}). This shared representation captures the underlying network state independent of the specific measurement modality.

\textbf{Indicator Decoders.} To ensure that the latent representation retains physical interpretability and does not discard information about individual indicators, we introduce symmetric indicator decoders $\psi_i: \mathbb{R}^d \rightarrow \mathbb{R}^{d_i}$ that reconstruct each indicator from its latent encoding:
$
\hat{x}^{(i)} = \psi_i(z^{(i)}).
$
These decoders act as regularizers, preventing the encoders from collapsing to trivial solutions and maintaining a faithful representation of observable network properties.

\textbf{Task-Specific Algorithms.} For each network tomography subtask $m \in \{\text{link}, \text{OD}, \text{topology}\}$, we apply a task-specific algorithm $\Gamma_m$ that takes the denoised reconstructed indicators as input:
$
\hat{y}_t = \Gamma_m(\{\hat{x}^{(i)}\}),
$
where $\{\hat{x}^{(i)}\}$ are the reconstructed network indicators from the decoders. For link-level estimation, $\Gamma_{\text{link}}$ estimates per-link performance metrics. For OD estimation, $\Gamma_{\text{OD}}$ produces the traffic matrix. For topology inference, $\Gamma_{\text{topology}}$ predicts the adjacency matrix. Note that these algorithms maybe are not differentiable, so task losses cannot backpropagate through the decoders, but gradients can still flow back to the encoders through the latent representations.

\underline{\textbf{\textit{Gradient in Shared Latent Space.}}}
After encoding different network indicators into the shared latent representation $z \in \mathbb{R}^d$, the subsequent multi-task optimization exhibits favorable gradient properties. Recall that task-specific decoders $\Gamma_t$ map from this shared space to different prediction targets. The following result explains why joint optimization in this shared latent space improves convergence.

\begin{proposition}[Gradient Reduction via Shared Subspace]
\label{prop: gradient_reduction}
Consider $m$ component gradients $g_1, \ldots, g_m \in \mathbb{R}^p$ with weight coefficients $\boldsymbol{\lambda} = (\lambda_1, \ldots, \lambda_m)^\top \in \mathbb{R}^m$. Define the composite gradient as
$g_{\mathrm{tot}} = \sum_{i=1}^m \lambda_i g_i = A\boldsymbol{\lambda},$
where $A = [g_1, \ldots, g_m] \in \mathbb{R}^{p \times m}$. Assume the conditions hold (detialed in Appendix~\ref{sec: appendix2}), the composite gradient satisfies
$$\|g_{\mathrm{tot}}\|_2^2 \leq (1 + \eta) \cdot \frac{1 + \delta}{r} \|\boldsymbol{\lambda}\|_2^2 \sum_{i=1}^m \|g_i\|_2^2,$$
where $\eta = O(\varepsilon\sqrt{r})$. When $\varepsilon, \delta$ are small and $r \ll m$, this provides a $1/r$ improvement over the naive bound.
\end{proposition}


\noindent
We use Proposition~\ref{prop: gradient_reduction} to reveal that by forcing all tasks to operate through the shared latent space $z$ (as enforced by our encoder-decoder architecture), the composite gradient magnitude is controlled by the effective subspace dimension $r$ rather than the number of tasks $m$ or the ambient latent dimension $d$.

\subsection{Learning Objective}

The training objective of \textsc{PlatoNT} comprises two main loss terms that jointly optimize representation alignment and reconstruction fidelity, with an optional task performance term.

\textbf{Alignment Loss.} To enforce multi-indicator alignment, we maximize the mutual information between latent representations of different indicators measured at the same time. Following the formulation in Eq.~(\ref{eq: app_mi}), we define the alignment loss as:
\begin{equation}
\mathcal{L}_{\text{align}} = -\frac{1}{N} \sum_{i \neq j} \sum_{n=1}^{N} \log \frac{\exp(s(z^{(i)}_{t_n}, z^{(j)}_{t_n}) / \tau)}{\frac{1}{N} \sum_{m=1}^{N} \exp(s(z^{(i)}_{t_n}, z^{(j)}_{t_m}) / \tau)},
\label{eq: align}
\end{equation}
where $s(\cdot, \cdot)$ is the cosine similarity, $\tau$ is the temperature parameter, and $N$ is the batch size. This contrastive objective encourages representations from the same time step to cluster together while pushing apart representations from different time steps, effectively aligning indicators that share a common network state.

\textbf{Reconstruction Loss.} To preserve the physical meaning of individual indicators and prevent information loss during encoding, we design a hybrid reconstruction objective that accounts for the noisy nature of network measurements.

In practice, network indicators obtained from probing flows are inherently noisy and may not accurately reflect true link properties. However, we assume access to a small set of high-quality ground-truth samples from well-monitored network segments. Let $\mathcal{D}_{\text{clean}}$ denote the clean sample set and $\mathcal{D}_{\text{noisy}}$ denote the noisy measurements. We define the reconstruction loss as:
\begin{equation}
\begin{aligned}
\mathcal{L}_{\text{rec}} & = \sum_{i} \bigl[\mathbb{E}_{x^{(i)} \in \mathcal{D}_{\text{clean}}} \bigl[ \| \tilde{x}^{(i)} - \psi_i(\phi_i(x^{(i)})) \|_2^2 \bigr] \\ 
& + \mathbb{E}_{x^{(i)} \in \mathcal{D}_{\text{noisy}}} \bigl[ \| x^{(i)} - \psi_i(\phi_i(x^{(i)})) \|_2^2 \bigr] \bigr],
\end{aligned}
\label{eq: rec}
\end{equation}

where $\tilde{x}^{(i)}$ represents the clean ground-truth for indicator $i$. 
The first term encourages the decoder to reconstruct clean network states when available, effectively learning to denoise the latent representation. The second term maintains fidelity to noisy observations when ground truth is unavailable, preventing the model from discarding information entirely.

\textbf{Task Loss (Optional).} When ground-truth labels for tomography tasks are available, we can optionally include task-specific losses to guide the learning of latent representations. For link-level estimation and OD traffic prediction, we use $\mathcal{L}_{m} = \mathbb{E}_{(z, y_{m})} [ \| \Gamma_{m}(\{\hat{x}^{(i)}\}) - y_{m} \|_2^2 ]$ where $m \in \{\text{link}, \text{OD}\}$, and for topology inference, we use $\mathcal{L}_{\text{topo}} = -\sum_{i,j} [ A_{ij} \log \hat{A}_{ij} + (1 - A_{ij}) \log (1 - \hat{A}_{ij}) ]$ where $A$ is the ground-truth adjacency matrix. The combined task loss is $\mathcal{L}_{\text{task}} = \sum_{t} \lambda_t \cdot \mathcal{L}_t$. Since some task algorithms are not differentiable, these losses provide gradients only to the encoders through the latent representations.

\begin{algorithm}[t]
\caption{\textsc{PLATONT} Training Algorithm}
\label{alg:platont}
\textbf{Input:} Network indicators $\{x^{(i)}\}_{i=1}^{M}$, clean samples $\mathcal{D}_{\text{clean}}$, noisy samples $\mathcal{D}_{\text{noisy}}$, task labels $\{y_t\}$ (optional) \\
\textbf{Output:} Encoders $\{\phi_i\}$, decoders $\{\psi_i\}$ \\
\textbf{Require:} Hyperparameters: $\lambda_1, \lambda_2, \lambda_3, \tau$ \\
\vspace{-0.5cm}
\begin{algorithmic}[1]
\ENSURE Trained model parameters
\STATE \textcolor{blue}{// \textbf{Phase 1:} Multi-indicator Encoding}
\FOR{each indicator type $i = 1, \ldots, M$}
    \STATE Encode to shared latent: $z^{(i)} = \phi_i(x^{(i)})$
\ENDFOR
\STATE Compute alignment loss $\mathcal{L}_{\text{align}}$ \COMMENT{Eq.~(\ref{eq: align})}

\STATE \textcolor{blue}{// \textbf{Phase 2:} Reconstruction with Denoising}
\FOR{each indicator type $i = 1, \ldots, M$}
    \STATE Reconstruct: $\hat{x}^{(i)} = \psi_i(z^{(i)})$
\ENDFOR
\STATE Compute hybrid reconstruction loss $\mathcal{L}_{\text{rec}}$ \COMMENT{Eq.~(\ref{eq: rec})}

\STATE \textcolor{blue}{// \textbf{Phase 3:} Task Execution (Optional)}
\IF{task labels available}
    \FOR{each task $t \in \{\text{link, OD, topology}\}$}
        \STATE Compute: $\hat{y}_t = \Gamma_t(\{\hat{x}^{(i)}\})$
    \ENDFOR
    \STATE Compute $\mathcal{L}_{\text{task}}$
\ENDIF

\STATE \textcolor{blue}{// \textbf{Phase 4:} Joint Optimization}
\STATE Compute total loss $\mathcal{L}_{\text{total}}$  \COMMENT{Eq.~(\ref{eq: total})}
\STATE Update parameters: $\theta \leftarrow \theta - \eta \nabla_\theta \mathcal{L}_{\text{total}}$

\RETURN Trained  $\{\phi_i\}$, $\{\psi_i\}$ \\
\end{algorithmic}
\end{algorithm}

\textbf{Unified Objective.} The complete training objective is:
\begin{equation}
\mathcal{L}_{\text{total}} = \lambda_1 \mathcal{L}_{\text{align}} + \lambda_2 \mathcal{L}_{\text{rec}} + \lambda_3 \mathcal{L}_{\text{task}}.
\label{eq: total}
\end{equation}
where $\lambda_1$, $\lambda_2$, and $\lambda_3$ are hyperparameters that balance the three objectives. The task loss term is optional and can be set to zero by using $\lambda_3 = 0$ when task labels are unavailable. In practice, we set $\lambda_1 = 1.0$ and $\lambda_2 = 2$, and $\lambda_3 = 1.0$ when using task supervision.

This unified objective realizes the design principles outlined earlier: $\mathcal{L}_{\text{align}}$ explicitly aligns multi-indicator representations, while $\mathcal{L}_{\text{task}}$ optionally regularizes the latent space through task supervision. The reconstruction loss $\mathcal{L}_{\text{rec}}$ serves as a bridge, ensuring that the learned representation remains grounded in observable network properties while supporting diverse inference tasks.

\section{Experiment}
\label{sec: exp}
We evaluate \textsc{PlatoNT} on three network tomography tasks: link-level parameter estimation, OD traffic matrix prediction, and topology inference. Experiments are conducted on multiple real-world network topologies under noisy environments, comparing against multi-view learning baselines.

\subsection{Experiment Settings}

\textbf{Simulation Environment and DataSet.} We evaluate \textsc{\textsc{PlatoNT}} on a publicly available dataset generated using OMNeT++~\cite{Varga2010OMNeT}, a widely-used discrete event packet-level network simulator. OMNeT++ provides high-fidelity simulation of network protocols and enables accurate measurement of network performance metrics at both link and path levels. The simulator captures the dynamic behavior of network traffic under various routing configurations and workload conditions, making it suitable for evaluating network tomography methods. Each simulation run is configured with specific topology, routing policy, and traffic matrix, allowing us to collect comprehensive network state observations across diverse operational scenarios.

\textbf{Topology Description.} To comprehensively evaluate the effectiveness of \textsc{\textsc{PlatoNT}}, we conduct experiments on six real-world network topologies in TopoHub~\cite{topohub}: AGIS, ChinaNet, GEANT, CANADA, Germany-17/50. These topologies represent different scales and structural characteristics commonly found in operational networks. The training dataset contains networks with 25-50 nodes, while the validation and test datasets include larger networks with 51-300 nodes to assess scalability and generalization. Each topology is associated with multiple routing configurations and traffic matrices, resulting in hundreds of distinct network states. The routing matrix $\mathbf{R} \in \{0,1\}^{|P| \times |L|}$ for each sample specifies the path-link incidence relations, where $|P|$ denotes the number of source-destination paths and $|L|$ denotes the number of links.

\textbf{Network Indicators.} The datasets\footnote{Datasets is  \href{https://bnn.upc.edu/challenge/gnnet2021/dataset/}{Dataset\_URL\_1} and \href{https://sndlib.put.poznan.pl/}{Dataset\_URL\_2}.} provides network performance metrics at multiple granularities. At the \underline{\textit{link level}}, we have per-link delay, loss rate, bandwidth utilization, and queue statistics. At the \underline{\textit{path level}}, the dataset includes end-to-end measurements for each source-destination pair: mean per-packet delay, jitter (delay variation), and packet loss rate. Additionally, the dataset contains \textit{\underline{origin-destination (OD)} traffic matrices} that specify the traffic volume between each source-destination pair, as well as port statistics such as queue size and utilization. For link-level parameter estimation, we focus on inferring per-link delay, loss, and bandwidth from path-level aggregated measurements. For OD traffic estimation, we aim to reconstruct the traffic matrix from link load observations. For topology inference, we predict the adjacency matrix from end-to-end delay measurements. This multi-indicator structure naturally aligns with our unified framework, where different metrics serve as complementary views of the underlying network state.

\subsection{Baselines}

We compare \textsc{\textsc{PlatoNT}} against representative multi-view learning baselines that align with our problem formulation.

\underline{P}rincipal \underline{C}omponent Analysis \textbf{(PCA).} As the most widely-used dimensionality reduction method, PCA serves as our fundamental baseline. We apply PCA to the concatenated multi-indicator features to obtain a low-dimensional representation. While PCA provides a simple and efficient approach, it does not explicitly model the statistical dependencies between different indicators, treating them as a single unified input rather than as complementary views of the same underlying network state.

\underline{C}anonical \underline{C}orrelation \underline{A}nalysis \textbf{(CCA)}~\cite{kumar2020multi}. CCA learns projection matrices that maximize the linear correlation between two views in a shared subspace. For network indicators $\mathbf{C}$ and $\mathbf{D}$, CCA finds projections $\mathbf{W}_c$ and $\mathbf{W}_d$ such that $\hat{\mathbf{C}} = \mathbf{C}\mathbf{W}_c$ and $\hat{\mathbf{D}} = \mathbf{D}\mathbf{W}_d$ are maximally correlated. Unlike our contrastive approach that aligns representations from the same network state while separating different states, CCA enforces strict linear correlation and requires different views to be mapped to orthogonal subspaces.

\underline{P}rojection \underline{M}atrix \underline{P}roduct \underline{D}ecomposition \textbf{(PMPD)}~\cite{sergazinov2025ppdd}. PMPD estimates signal and noise subspaces for multi-view data using truncated SVD with Gavish-Donoho thresholding, combined with rotational Bootstrap for subspace alignment. It leverages random matrix theory to filter noise directions and determines the joint rank by counting significant singular values of the projection matrix product. While PMPD offers a principled subspace decomposition approach, it assumes linear subspace structures and does not incorporate temporal co-occurrence information. In contrast, \textsc{\textsc{PlatoNT}} explicitly models temporal alignment through contrastive learning and supports nonlinear transformations via neural network.

For fairness, in different NT subtasks we adopt standard task-specific baseline algorithms: Range tomography~\cite{range_nt} for link performance inference, an established OD estimation method~\cite{erramill2006independent} for OD traffic recovery, and the RNJ~\cite{ni2009efficient} for topology inference. In the bias distribution analysis, measurement noise levels of $0.05$, $0.1$, and $0.2$ are injected, and the reported results are averaged over these three settings.

 \begin{figure}[t]
	\centering
	\includegraphics[width=0.85\linewidth]{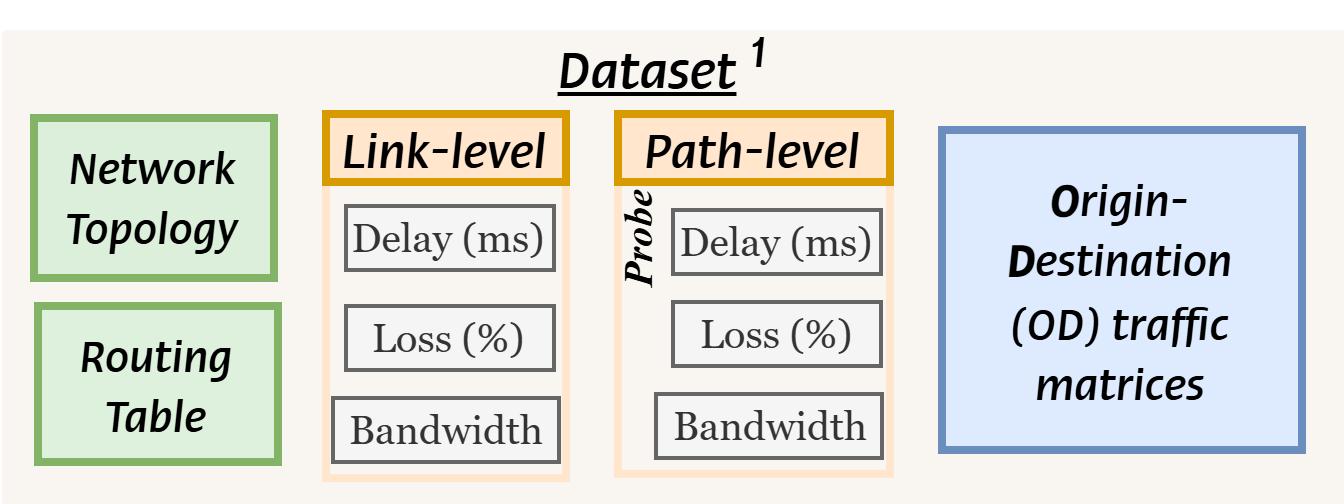}
     \caption{Overview of dataset structure.} 
     \vspace{-0.7cm}
	\label{fig: datasets}
\end{figure}

\subsection{Bias Distribution Analysis}

To validate that \textsc{PlatoNT}'s can accurately reconstruct individual network indicators, we examine the prediction bias and error distributions across delay, loss, and bandwidth metrics under noisy measurement conditions.

\begin{figure*}[t]
	\centering
	\includegraphics[width=\textwidth]{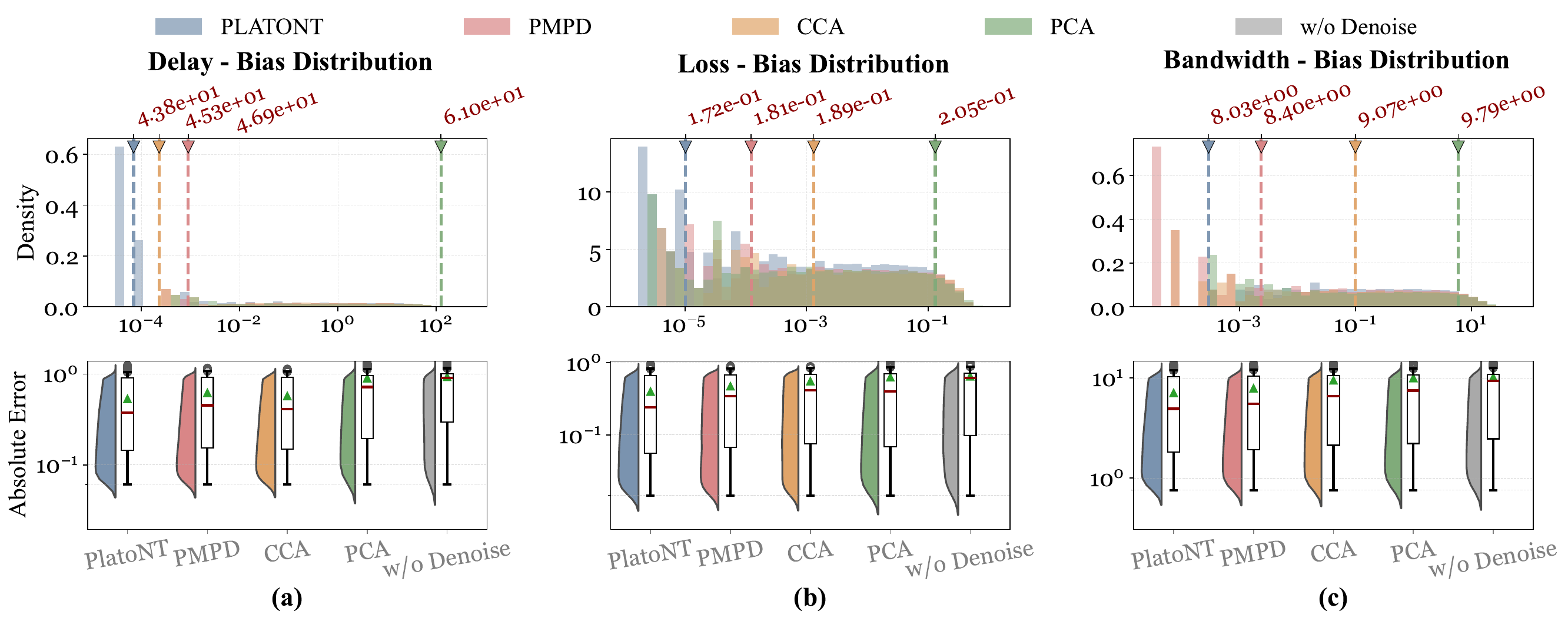}
     \caption{Bias distribution and absolute error comparison across network indicators under noisy environments. \textbf{Top row}: probability density distributions of prediction bias for (a) delay, (b) loss, and (c) bandwidth, each employing \textbf{dual x-axes} to depict values on both linear and logarithmic scales; dashed lines indicate mean values. \textbf{Bottom row}: box plots of absolute errors across different network links over multiple runs. Results are averaged over different noise conditions.}
     \vspace{-0.5cm}
	\label{fig: indictors}
\end{figure*}
\begin{table}[t]
\centering
\caption{Performance comparison of congested link diagnosis across different network topologies under noisy environments. Colored percentages show relative loss ({\color{red} $\blacktriangledown$}) or degradation ({\color{darkgreen} $\blacktriangle$}) compared to noise-free baselines.}
\label{tab: performance_link}
\resizebox{\linewidth}{!}{
\begin{tabular}{c|c|cccc}
\toprule
\textbf{Topo.} & \textbf{Algorithm} & \textbf{Precision} & \textbf{Recall} & \textbf{F1-Score} & \textbf{FPR} \\
\midrule
\multirow{5}{*}{\rotatebox[origin=c]{90}{CHINANET}} 
& \cellcolor{gray!20}w/o Noise & $\cellcolor{gray!20}0.955$ & $\cellcolor{gray!20}0.668$ & $\cellcolor{gray!20}0.745$ & $\cellcolor{gray!20}0.006$ \\
& PCA & $0.764_{\color{red}\blacktriangledown 20.01\%}$ & $0.618_{\color{red}\blacktriangledown 7.48\%}$ & $0.644_{\color{red}\blacktriangledown 13.59\%}$ & $0.067_{\color{darkgreen}\blacktriangle 1062.80\%}$ \\
& CCA & $0.856_{\color{red}\blacktriangledown 10.33\%}$ & $0.644_{\color{red}\blacktriangledown 3.55\%}$ & $0.694_{\color{red}\blacktriangledown 6.81\%}$ & $0.038_{\color{darkgreen}\blacktriangle 562.92\%}$ \\
& PMPD & $0.889_{\color{red}\blacktriangledown 6.92\%}$ & $0.648_{\color{red}\blacktriangledown 2.94\%}$ & $0.709_{\color{red}\blacktriangledown 4.91\%}$ & $0.027_{\color{darkgreen}\blacktriangle 376.42\%}$ \\
& \textbf{\textsc{PlantoNT}} & $\mathbf{0.915_{\color{red}\blacktriangledown 4.21\%}}$ & $\mathbf{0.660_{\color{red}\blacktriangledown 1.13\%}}$ & $\mathbf{0.724_{\color{red}\blacktriangledown 2.80\%}}$ & $\mathbf{0.019_{\color{darkgreen}\blacktriangle 234.72\%}}$ \\
\midrule
\multirow{5}{*}{\rotatebox[origin=c]{90}{AGIS}} 
& \cellcolor{gray!20}w/o Noise & $\cellcolor{gray!20}0.952$ & $\cellcolor{gray!20}0.761$ & $\cellcolor{gray!20}0.813$ & $\cellcolor{gray!20}0.013$ \\
& PCA & $0.778_{\color{red}\blacktriangledown 18.24\%}$ & $0.710_{\color{red}\blacktriangledown 6.71\%}$ & $0.714_{\color{red}\blacktriangledown 12.09\%}$ & $0.070_{\color{darkgreen}\blacktriangle 421.22\%}$ \\
& CCA & $0.870_{\color{red}\blacktriangledown 8.60\%}$ & $0.740_{\color{red}\blacktriangledown 2.66\%}$ & $0.766_{\color{red}\blacktriangledown 5.80\%}$ & $0.038_{\color{darkgreen}\blacktriangle 186.59\%}$ \\
& PMPD & $0.896_{\color{red}\blacktriangledown 5.86\%}$ & $0.748_{\color{red}\blacktriangledown 1.70\%}$ & $0.784_{\color{red}\blacktriangledown 3.54\%}$ & $0.029_{\color{darkgreen}\blacktriangle 121.25\%}$ \\
& \textbf{\textsc{PlantoNT}} & $\mathbf{0.919_{\color{red}\blacktriangledown 3.46\%}}$ & $\mathbf{0.751_{\color{red}\blacktriangledown 1.24\%}}$ & $\mathbf{0.796_{\color{red}\blacktriangledown 2.03\%}}$ & $\mathbf{0.026_{\color{darkgreen}\blacktriangle 93.33\%}}$ \\
\midrule
\multirow{5}{*}{\rotatebox[origin=c]{90}{GEANT}} 
& \cellcolor{gray!20}w/o Noise & $\cellcolor{gray!20}0.952$ & $\cellcolor{gray!20}0.781$ & $\cellcolor{gray!20}0.825$ & $\cellcolor{gray!20}0.015$ \\
& PCA & $0.800_{\color{red}\blacktriangledown 15.99\%}$ & $0.731_{\color{red}\blacktriangledown 6.31\%}$ & $0.732_{\color{red}\blacktriangledown 11.31\%}$ & $0.073_{\color{darkgreen}\blacktriangle 389.68\%}$ \\
& CCA & $0.879_{\color{red}\blacktriangledown 7.63\%}$ & $0.770_{\color{red}\blacktriangledown 1.35\%}$ & $0.789_{\color{red}\blacktriangledown 4.32\%}$ & $0.040_{\color{darkgreen}\blacktriangle 169.08\%}$ \\
& PMPD & $0.907_{\color{red}\blacktriangledown 4.72\%}$ & $0.765_{\color{red}\blacktriangledown 1.98\%}$ & $0.796_{\color{red}\blacktriangledown 3.52\%}$ & $0.033_{\color{darkgreen}\blacktriangle 123.27\%}$ \\
& \textbf{\textsc{PlantoNT}} & $\mathbf{0.924_{\color{red}\blacktriangledown 2.99\%}}$ & $\mathbf{0.772_{\color{red}\blacktriangledown 1.04\%}}$ & $\mathbf{0.807_{\color{red}\blacktriangledown 2.17\%}}$ & $\mathbf{0.026_{\color{darkgreen}\blacktriangle 78.56\%}}$ \\
\midrule
\multirow{5}{*}{\rotatebox[origin=c]{90}{CANADA}} 
& \cellcolor{gray!20}w/o Noise & $\cellcolor{gray!20}0.953$ & $\cellcolor{gray!20}0.793$ & $\cellcolor{gray!20}0.840$ & $\cellcolor{gray!20}0.011$ \\
& PCA & $0.819_{\color{red}\blacktriangledown 14.05\%}$ & $0.735_{\color{red}\blacktriangledown 7.29\%}$ & $0.763_{\color{red}\blacktriangledown 9.13\%}$ & $0.045_{\color{darkgreen}\blacktriangle 312.88\%}$ \\
& CCA & $0.889_{\color{red}\blacktriangledown 6.74\%}$ & $0.761_{\color{red}\blacktriangledown 3.97\%}$ & $0.798_{\color{red}\blacktriangledown 4.98\%}$ & $0.029_{\color{darkgreen}\blacktriangle 161.17\%}$ \\
& PMPD & $0.914_{\color{red}\blacktriangledown 4.12\%}$ & $0.777_{\color{red}\blacktriangledown 1.93\%}$ & $0.815_{\color{red}\blacktriangledown 2.92\%}$ & $0.023_{\color{darkgreen}\blacktriangle 113.81\%}$ \\
& \textbf{\textsc{PlantoNT}} & $\mathbf{0.927_{\color{red}\blacktriangledown 2.72\%}}$ & $\mathbf{0.772_{\color{red}\blacktriangledown 2.58\%}}$ & $\mathbf{0.816_{\color{red}\blacktriangledown 2.82\%}}$ & $\mathbf{0.018_{\color{darkgreen}\blacktriangle 62.04\%}}$ \\
\bottomrule
\end{tabular}
}
\vspace{-0.4cm}
\end{table}

We analyze the bias distribution between predicted network indicators and ground-truth values under noisy environments, as shown in Figure~\ref{fig: indictors}. Performance is evaluated under both channel noise and random noise, with results averaged across these two conditions. The top row presents probability density distributions of prediction bias for delay (a), loss (b), and bandwidth (c) estimation, with vertical dashed lines indicating mean bias for each method. \textsc{PlatoNT} achieves the smallest mean bias across all three indicators, with values of $4.38\times10^{-1}$ for delay, $1.72\times10^{-1}$ for loss, and $8.03\times10^{0}$ for bandwidth. Compared to baseline methods, \textsc{PlatoNT} reduces mean bias by 6.8\%, 5.0\%, and 3.8\% relative to PMPD, and by 28.2\%, 9.9\%, and 11.3\% relative to the weakest baseline PCA. Other multi-view approaches such as CCA and PCA show higher bias values, with CCA achieving $4.53\times10^{-1}$, $1.89\times10^{-1}$, and $9.07\times10^{0}$, while PCA reaches $6.10\times10^{-1}$, $2.05\times10^{-1}$, and $9.79\times10^{0}$ for the three indicators respectively. These results demonstrate that \textsc{PlatoNT}'s multi-indicator alignment effectively reduces prediction bias in noisy network environments.

The corresponding box plots in the bottom row reveal absolute error distributions across different network links over multiple runs. \textsc{PlatoNT} consistently exhibits the tightest error distributions with the bulk of predictions concentrated near zero across all three network indicators. The compact interquartile ranges demonstrate that \textsc{PlatoNT} maintains stable and accurate predictions, with significantly fewer outliers compared to baseline methods. In contrast, baseline methods show wider error spreads with more frequent extreme values, particularly for PCA and the ablation variant w/o Denoise. The concentrated distribution near zero indicates that \textsc{PlatoNT}'s unified latent representation effectively captures the underlying network state, leading to more consistent predictions across diverse network conditions and measurement scenarios.

\subsection{Link Performance Estimation}

Link-level performance estimation encompasses various tasks such as delay inference, loss rate estimation, and bandwidth prediction. We select congested link identification as a representative evaluation metric, as it is a classical and widely-studied problem in network tomography that directly impacts network management decisions.

\begin{figure*}[t]
	\centering
	\includegraphics[width=0.95\textwidth]{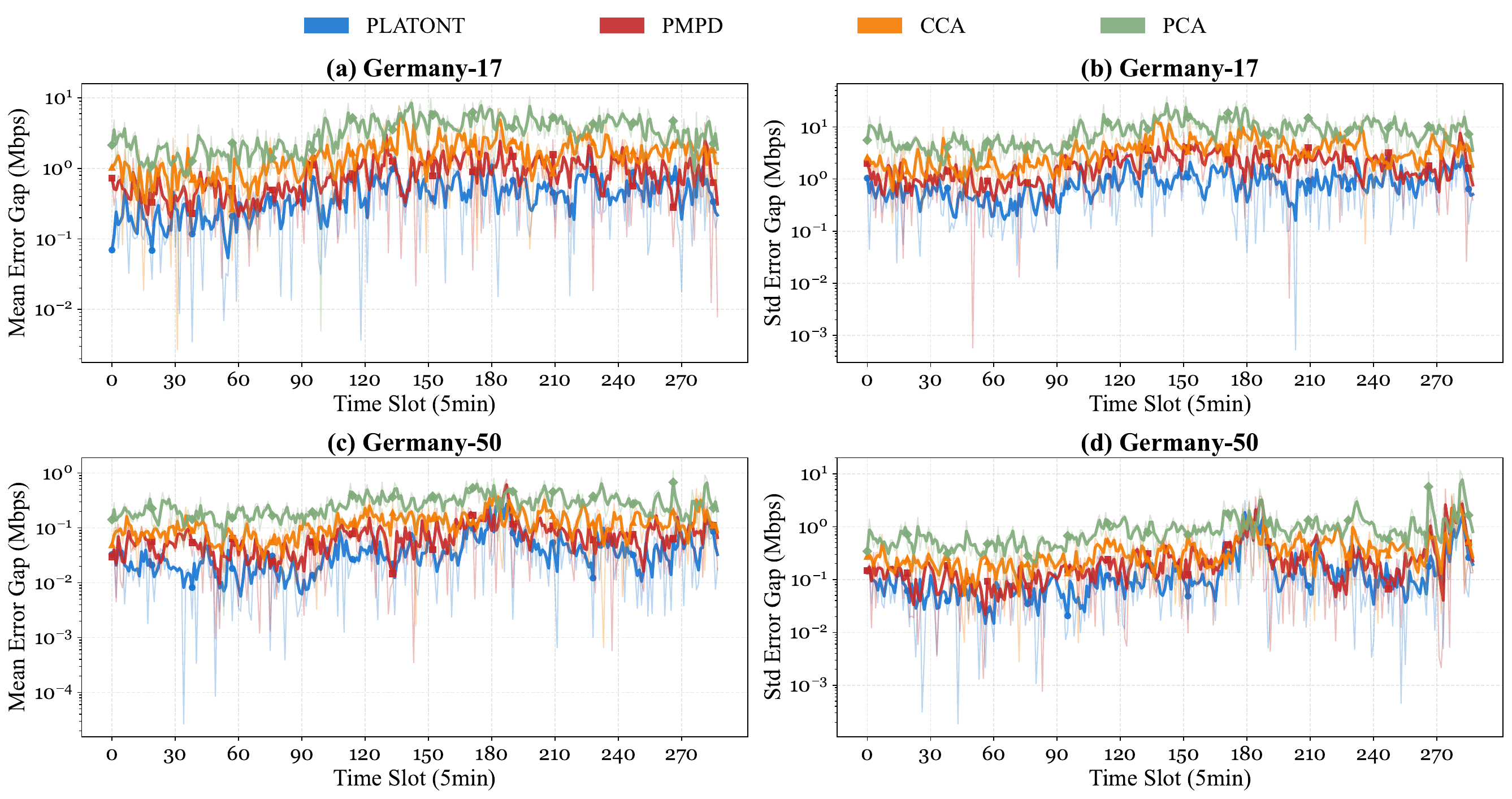}
     \caption{OD traffic matrix estimation performance over time. Rows show results for Germany-17 (a-b) and Germany-50 (c-d) topologies. \textbf{Left column}: mean error gap over 300 time slots (5-minute intervals). \textbf{Right column}: standard deviation of error gap. Error gap is measured as deviation from ground-truth OD matrix obtained using noise-free indicators.}
    \vspace{-0.5cm}
	\label{fig: od_performance}
\end{figure*}

 \begin{figure}[t]
	\centering
	\includegraphics[width=\linewidth]{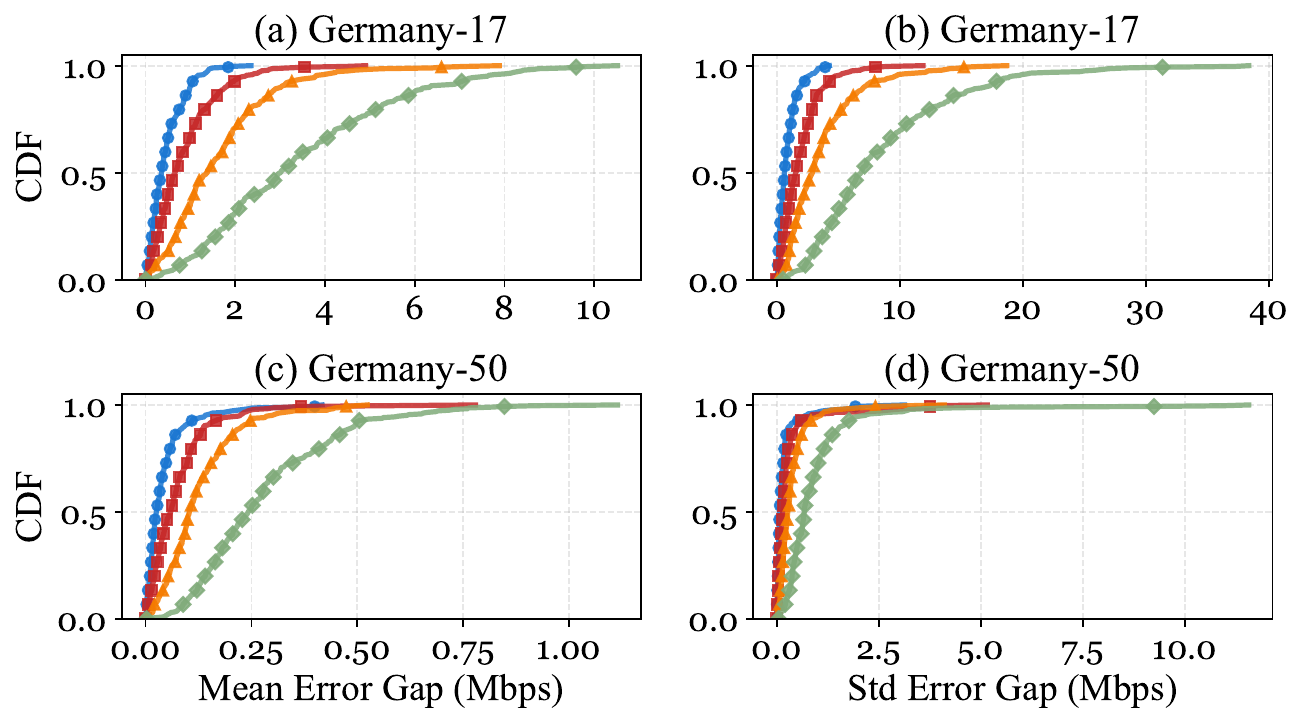}
     \vspace{-0.3cm}
     \caption{CDFs of OD prediction errors. Rows show Germany-17 (a-b) and Germany-50 (c-d) topologies. \textbf{Left column}: CDF of mean error gap. \textbf{Right column}: CDF of standard deviation error gap.}
     \vspace{-0.5cm}
	\label{fig: od_cdf}
\end{figure}

The performance comparison across four real-world network topologies is presented in Table~\ref{tab: performance_link}, where we evaluate congested link diagnosis using Precision, Recall, F1-Score, and False Positive Rate (FPR). Gray-shaded rows indicate the noise-free upper bound performance. \textsc{PlatoNT} consistently outperforms all baseline methods across all topologies and metrics. On the CHINANET topology, \textsc{PlatoNT} achieves a Precision of 0.915, Recall of 0.660, and F1-Score of 0.724, with only 4.21\%, 1.13\%, and 2.80\% degradation compared to the noise-free scenario, while maintaining the lowest FPR of 0.019. In contrast, traditional dimensionality reduction methods PCA and CCA suffer from substantial performance drops of 20.01\% and 10.33\% in Precision respectively, with FPR increasing by over 500\%. The baseline PMPD shows improved robustness with 6.92\% Precision degradation, yet \textsc{PlatoNT} still surpasses it by 2.92\% in Precision and reduces FPR by 29.63\%.

Across other topologies, \textsc{PlatoNT} maintains competitive performance. On AGIS, \textsc{PlatoNT} achieves 0.919 Precision, and 0.796 F1-Score, outperforming PMPD by 2.57\%, and 1.53\% respectively. On GEANT and CANADA topologies, \textsc{PlatoNT} exhibits smaller performance gaps relative to noise-free conditions, with F1-Score degradations of only 2.17\% and 2.82\%, compared to PMPD's 3.52\% and 2.92\%. \textsc{PlatoNT} achieves FPR reductions of 21.21\%, 10.34\%, and 21.74\% compared to PMPD on AGIS, GEANT, and CANADA respectively. These results across diverse network topologies indicate that \textsc{PlatoNT}'s unified latent representation captures shared network state information effectively, enabling more accurate congested link identification under noisy measurement conditions.

\subsection{Origin-Destination Traffic Matrix Estimation}

To assess the temporal stability and scalability of \textsc{PlatoNT}'s denoised representations, we evaluate OD traffic matrix estimation performance across different network scales and over extended time periods.

\begin{figure*}[t]
	\centering
	\includegraphics[width=0.99\textwidth]{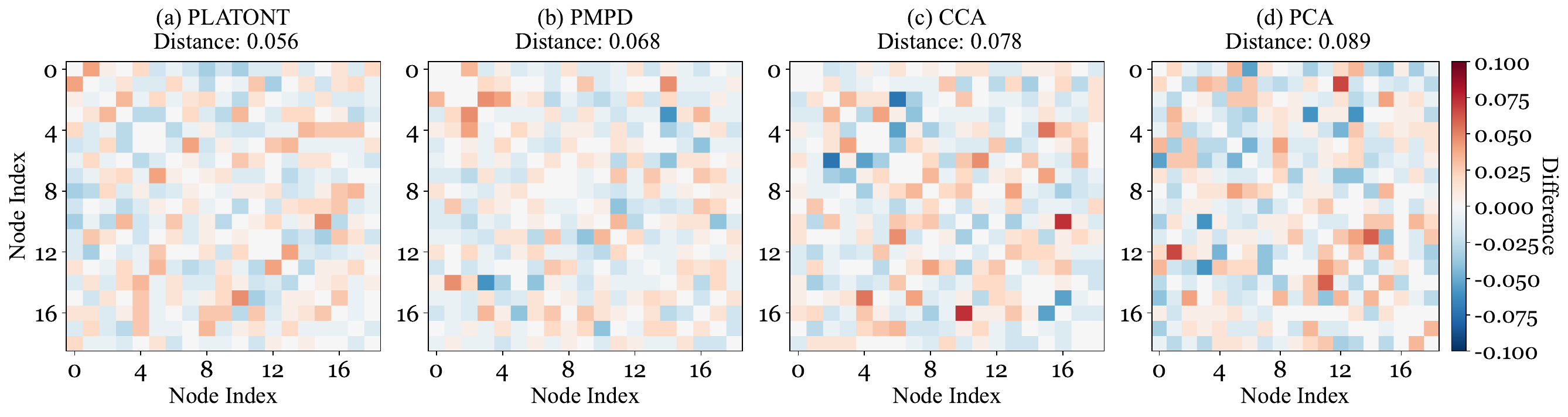}
     \caption{Topology inference visualization on a 19-node tree topology. Heatmaps show the Hamming difference between inferred and ground-truth adjacency matrices for (a) \textsc{PlatoNT}, (b) PMPD, (c) CCA, and (d) PCA. Color intensity indicates deviation magnitude (blue: negative, red: positive). Distance values measure overall reconstruction error. }
     \vspace{-0.5cm}
	\label{fig: topo_heatmap}
\end{figure*}

The temporal performance is illustrated in Figure~\ref{fig: od_performance}, showing results for Germany-17 (top row, a-b) and Germany-50 (bottom row, c-d). The left column displays mean error gap while the right column shows standard deviation error gap over 300 time slots (5-minute intervals), where error gap measures deviation from the ground-truth OD matrix obtained using noise-free indicators. On the Germany-17 topology, \textsc{PlatoNT} maintains the lowest mean error gap with values consistently around $10^{-1}$ Mbps, compared to PMPD ($10^{0}$ Mbps), CCA ($10^{0}$ Mbps), and PCA ($10^{1}$ Mbps). The standard deviation error gap shows similar patterns, with \textsc{PlatoNT} achieving the smallest fluctuations across all time slots. On the Germany-50 topology with larger network scale, \textsc{PlatoNT} demonstrates more pronounced advantages, with mean error gaps approaching the $10^{-2}$ Mbps scale while baseline methods range from $10^{-1}$ to $10^{0}$ Mbps. The temporal stability of \textsc{PlatoNT}'s predictions, evidenced by consistent performance across 300 time slots, indicates robust generalization to dynamic traffic patterns.

To better understand the error distribution characteristics, we present CDFs in Figure~\ref{fig: od_cdf} for both mean error gap and standard deviation error gap across both topologies. For the Germany-17 topology, \textsc{PlatoNT} achieves 80\% of predictions with mean error gap below 1 Mbps, while PMPD, CCA, and PCA require error thresholds of 2 Mbps, 3 Mbps, and 6 Mbps respectively to reach the same cumulative probability. The CDF curves show that \textsc{PlatoNT}'s error distribution is heavily concentrated in the low-error region, with over 90\% of predictions having mean error gaps below 1.5 Mbps. For standard deviation error gap on Germany-17, \textsc{PlatoNT} achieves the steepest CDF curve, with 80\% of predictions below 5 Mbps compared to 10 Mbps for PMPD and over 20 Mbps for PCA. Similar trends are observed on the Germany-50 topology, where \textsc{PlatoNT}'s CDF curves consistently dominate all baselines across both error metrics. The superior CDF characteristics demonstrate that \textsc{PlatoNT}'s unified latent representation not only reduces average prediction errors but also improves prediction consistency across different OD pairs and time periods.

\begin{table}[t]
\centering
\caption{Route Table inference performance comparison across different real-world network topologies under noisy conditions. Results are reported as mean $\pm$ standard deviation over multiple runs.}
\label{tab: topo_performance}
\resizebox{\linewidth}{!}{
\begin{tabular}{c|c|cc}
\toprule
\textbf{Topology} & \textbf{ \quad Algorithm \quad} & \textbf{Hamming Distance} & \textbf{Frobenius Distance} \\
\midrule
\multirow{4}{*}{AGIS} 
& PCA & $0.075_{\pm 0.045}$ & $2.831_{\pm 1.699}$ \\
& CCA & $0.053_{\pm 0.046}$ & $2.207_{\pm 1.671}$ \\
& PMPD & $0.055_{\pm 0.051}$ & $2.177_{\pm 1.733}$ \\
& \cellcolor{blue!5}\textbf{\textsc{PlantoNT}} & $\cellcolor{blue!5}\mathbf{0.026_{\pm 0.040}}$ & $\cellcolor{blue!5}\mathbf{1.249_{\pm 1.483}}$ \\
\midrule
\multirow{4}{*}{CHINANET} 
& PCA & $0.038_{\pm 0.019}$ & $4.223_{\pm 1.588}$ \\
& CCA & $0.043_{\pm 0.018}$ & $4.322_{\pm 1.489}$ \\
& PMPD & $0.037_{\pm 0.020}$ & $3.783_{\pm 1.671}$ \\
& \cellcolor{blue!5}\textbf{\textsc{PlantoNT}} & $\cellcolor{blue!5}\mathbf{0.029_{\pm 0.024}}$ & $\cellcolor{blue!5}\mathbf{3.170_{\pm 2.105}}$ \\
\midrule
\multirow{4}{*}{CANADA} 
& PCA & $0.062_{\pm 0.033}$ & $6.918_{\pm 2.442}$ \\
& CCA & $0.057_{\pm 0.032}$ & $5.991_{\pm 2.650}$ \\
& PMPD & $0.051_{\pm 0.031}$ & $5.536_{\pm 2.884}$ \\
& \cellcolor{blue!5}\textbf{\textsc{PlantoNT}} & $\cellcolor{blue!5}\mathbf{0.030_{\pm 0.029}}$ & $\cellcolor{blue!5}\mathbf{3.766_{\pm 3.092}}$ \\
\midrule
\multirow{4}{*}{GEANT} 
& PCA & $0.048_{\pm 0.035}$ & $2.789_{\pm 1.600}$ \\
& CCA & $0.038_{\pm 0.035}$ & $2.336_{\pm 1.722}$ \\
& PMPD & $0.036_{\pm 0.032}$ & $2.148_{\pm 1.633}$ \\
& \cellcolor{blue!5}\textbf{\textsc{PlantoNT}} & $\cellcolor{blue!5}\mathbf{0.019_{\pm 0.030}}$ & $\cellcolor{blue!5}\mathbf{1.152_{\pm 1.629}}$ \\
\bottomrule
\end{tabular}
}
\vspace{-0.5cm}
\end{table}
\subsection{Network Topology Inference}

We examine whether \textsc{PlatoNT}'s multi-indicator alignment preserves structural information necessary for topology inference, testing the framework's ability to recover network connectivity from denoised end-to-end measurements.

Visual comparison of topology reconstruction quality is provided in Figure~\ref{fig: topo_heatmap} for a 19-node tree topology, where heatmaps represent the hamming difference between inferred adjacency matrices and ground truth. Each cell $(i,j)$ indicates the deviation in predicted connection strength between node $i$ and node $j$, with blue and red colors representing negative and positive differences respectively. \textsc{PlatoNT} achieves the smallest overall distance of 0.056, demonstrating the most accurate topology reconstruction with minimal structural deviations from the ground truth. The heatmap shows that \textsc{PlatoNT}'s predictions exhibit lighter color intensities and more uniform patterns, indicating smaller and more evenly distributed errors across all node pairs. In comparison, PMPD achieves a distance of 0.068, CCA reaches 0.078, and PCA performs worst with a distance of 0.089, representing 17.6\%, 28.2\%, and 37.1\% higher reconstruction errors compared to \textsc{PlatoNT}. The visual patterns reveal that baseline methods produce more concentrated error regions (darker patches) and higher variance in prediction differences, while \textsc{PlatoNT} maintains consistent accuracy across the entire adjacency matrix.

Quantitative results across real-world topologies route table are summarized in Table~\ref{tab: topo_performance}, using Hamming distance and Frobenius distance metrics. \textsc{PlatoNT} consistently achieves the lowest distances across all topologies. On the AGIS topology, \textsc{PlatoNT} attains a Hamming distance of $0.026$ and Frobenius distance of $1.249$, outperforming the second-best baseline PMPD by 52.7\% and 42.6\% respectively. On CHINANET, \textsc{PlatoNT} achieves $0.029$ Hamming distance and $3.170$ Frobenius distance, representing improvements of 21.6\% and 16.2\% over PMPD. More significant gains are observed on CANADA topology, where \textsc{PlatoNT} reduces Hamming distance to $0.030$ and Frobenius distance to $3.766$, achieving 41.2\% and 32.0\% improvements over PMPD. On GEANT, \textsc{PlatoNT} reaches $0.019$ and $1.152$ for the two metrics, demonstrating 47.2\% and 46.4\% reductions compared to PMPD. The consistently lower standard deviations of \textsc{PlatoNT} across most cases indicate more stable topology inference performance under varying noise conditions.

\section{Conclusion}
\label{sec: conclusion}
We introduced \textsc{PlatoNT}, a unified framework for network tomography based on the Platonic Representation Hypothesis. By treating different network indicators as projections of a shared latent network state, \textsc{PlatoNT} addresses the key limitation of traditional methods that process each indicator in isolation. Our approach combines multi-indicator alignment through contrastive learning and multi-task regularization by jointly optimizing link estimation, OD traffic prediction, and topology inference. We provide theoretical justification showing that contrastive learning exactly represents the PMI kernel (Theorem~\ref{thm: pmi_kernel}) and that joint optimization in a shared subspace reduces gradient magnitude by a factor of $1/r$ (Proposition~\ref{prop: gradient_reduction}). Experiments on multiple real-world network topologies demonstrate that \textsc{PlatoNT} consistently outperforms existing methods across all three tomography tasks, achieving better accuracy, robustness to noise, and generalization to unseen network conditions.

\appendices
\section{Proof of Theorem~\ref{thm: pmi_kernel}}
\label{sec: appendix1}
\begin{Proof}
Let $K \in \mathbb{R}^{N \times N}$ be the symmetric matrix defined by PMI over events $\{z_1, \ldots, z_N\}$ with entries
$$K_{ij} = K_{\mathrm{PMI}}(z_i, z_j) = \log \frac{P_{\mathrm{coor}}(z_i, z_j)}{P_{\mathrm{coor}}(z_i) P_{\mathrm{coor}}(z_j)}.$$

Assume there exists a parameter $\varepsilon \geq 0$ and $\rho_{\min} \in (0,1]$, and following below two assumptions~\cite{PRH_2024}:

\noindent
\textbf{(A1) (Off-diagonal smoothness):} 
$$K_{ij} \in [\log \rho_{\min},\, \log \rho_{\min} + \varepsilon], \text{ for all } i \neq j$$

\noindent
\textbf{(A2) (Diagonal lower bound):}
$$K_{ii} \geq N\varepsilon + \log \rho_{\min}, \text{ for all } i $$

\noindent
Define $\alpha := \max_{i \neq j} |K_{ij}|$ as the upper bound of off-diagonal absolute values. We prove the following two statements:
\noindent
\textbf{({\color{red} I}) General case:} If we choose the shift constant
$$C \geq \max\left(0,\, (N-1)\alpha - \min_i K_{ii}\right),$$
then there exists a mapping $f_X: \mathcal{X} \to \mathbb{R}^d$ such that $\langle f_X(x_i), f_X(x_j) \rangle = K_{ij} + C$ (i.e., $K + CI \succeq 0$).
\noindent
\textbf{({\color{red} II}) Under assumptions (A1) and (A2):} If additionally
$\varepsilon \geq N|\log \rho_{\min}|,$
then $K \succeq 0$ (i.e., $C = 0$ suffices), and there exists $f_X$ such that $\langle f_X(x_i), f_X(x_j) \rangle = K_{ij}$.

\noindent
{\color{red} \textbf{(I).}}
We decompose $K$ into diagonal and off-diagonal parts:
\begin{equation}
    K = D + R, \ D = \operatorname{diag}(K_{11}, \ldots, K_{NN}), \ R = K - D.
\end{equation}
\noindent
By Weyl's inequality~\cite{franklin2000matrix} for symmetric matrices:
\begin{equation}
\label{eq: weyl}
\lambda_{\min}(K) \geq \lambda_{\min}(D) - \|R\|_2 = \min_i K_{ii} - \|R\|_2.
\end{equation}

\noindent
For the symmetric matrix $R$, the spectral norm is bounded by the row-sum norm:
$$\|R\|_2 \leq \|R\|_\infty = \max_i \sum_{j \neq i} |K_{ij}| \leq (N-1)\alpha,$$

where $\alpha := \max_{i \neq j} |K_{ij}|$.
From \eqref{eq: weyl} and the above bound, we obtain the general lower bound
$$\lambda_{\min}(K) \geq \min_i K_{ii} - (N-1)\alpha.$$

\noindent
Therefore, if the right-hand side is non-negative, then $K \succeq 0$; if negative, a shift $C \geq -\lambda_{\min}(K)$ ensures $K + CI \succeq 0$. Substituting the upper bound yields the explicit sufficient shift
$$C \geq (N-1)\alpha - \min_i K_{ii},$$
which establishes \textbf{(I)}.

\noindent
{\color{red} \textbf{(II).}} Under assumption (A1), all off-diagonal entries satisfy $K_{ij} \in [\log \rho_{\min},\, \log \rho_{\min} + \varepsilon]$. Since $\rho_{\min} \in (0,1]$, we have $\log \rho_{\min} \leq 0$. Thus, taking the absolute value upper bound:
\begin{align} 
  \alpha & = \max_{i \neq j} |K_{ij}|  \leq \max\left(|\log \rho_{\min}|,\, |\log \rho_{\min} + \varepsilon|\right)  \nonumber \\  
  & = -\log \rho_{\min} + \varepsilon.   \nonumber
\end{align}

Using (A2), we have $\min_i K_{ii} \geq N\varepsilon + \log \rho_{\min}$. Substituting these bounds into the spectral lower bound:
\begin{align*}
\lambda_{\min}(&K) 
\geq \min_i K_{ii} - (N-1)\alpha \\
&\geq (N\varepsilon + \log \rho_{\min}) - (N-1)(-\log \rho_{\min} + \varepsilon) \\
&= N\varepsilon + \log \rho_{\min} - (N-1)(-\log \rho_{\min}) - (N-1)\varepsilon \\
&= \varepsilon + N\log \rho_{\min} 
\end{align*}

Therefore, when $\varepsilon \geq N|\log \rho_{\min}|$, the right-hand side is non-negative, yielding $\lambda_{\min}(K) \geq 0$, i.e., $K \succeq 0$. This completes the proof of \textbf{(II)}. $\hfill\square$
\end{Proof}

\vspace{-0.5cm}
\section*{Proof of Proposition~\ref{prop: gradient_reduction}}
\label{sec: appendix2}
\begin{Proof}
\noindent
We make the gradient structure precise through the following assumptions:

\noindent
\textbf{(S1)} There exists a linear subspace $U \subset \mathbb{R}^p$ with $\dim(U) = r \ll p$ such that for each $i$,
$$\|(I - P_U)g_i\|_2 \leq \varepsilon \|g_i\|_2,$$
where $P_U$ denotes the orthogonal projection onto $U$ and $\varepsilon \geq 0$ quantifies the residual energy outside $U$.

\noindent
\textbf{(S2)} Let $\tilde{A} = P_U A = [\tilde{g}_1, \ldots, \tilde{g}_m]$ be the projection of $A$ onto $U$. The Gram matrix $\tilde{A}\tilde{A}^\top$ satisfies
$$\lambda_{\max}(\tilde{A}\tilde{A}^\top) \leq (1 + \delta) \frac{\operatorname{trace}(\tilde{A}\tilde{A}^\top)}{r},$$
where $\delta \geq 0$ measures the deviation from spectral uniformity.

\noindent
Let $\tilde{A} = P_U A = [\tilde{g}_1, \ldots, \tilde{g}_m]$ denote the projection of $A$ onto the subspace $U$, and $E = (I - P_U)A = [e_1, \ldots, e_m]$ the residual component, so that $A = \tilde{A} + E$. The composite gradient decomposes as $g_{\mathrm{tot}} = A\boldsymbol{\lambda} = \tilde{A}\boldsymbol{\lambda} + E\boldsymbol{\lambda}$. We denote $\tilde{g}_{\mathrm{tot}} = \tilde{A}\boldsymbol{\lambda}$ and $e_{\mathrm{tot}} = E\boldsymbol{\lambda}$ as the subspace and residual components respectively. For notational convenience, define $S = \sum_{i=1}^m \|g_i\|_2^2$ as the total squared norm of all component gradients.

By the triangle inequality and squaring, we have
\begin{equation}
\label{eq:triangle}
\begin{aligned}
     \|g_{\mathrm{tot}}\|_2^2 & = \|\tilde{g}_{\mathrm{tot}} + e_{\mathrm{tot}}\|_2^2 \leq \Big(\|\tilde{g}_{\mathrm{tot}}\|_2 + \|e_{\mathrm{tot}}\|_2   \Big)^2 
  \\ &= \|\tilde{g}_{\mathrm{tot}}\|_2^2 + 2\|\tilde{g}_{\mathrm{tot}}\|_2 \|e_{\mathrm{tot}}\|_2 + \|e_{\mathrm{tot}}\|_2^2.  
\end{aligned}
\end{equation}
We now bound the residual component $\|e_{\mathrm{tot}}\|_2$ and the subspace component $\|\tilde{g}_{\mathrm{tot}}\|_2$ separately using the structural assumptions (S1) and (S2).

First, we bound the residual component. From assumption (S1), each residual satisfies $\|e_i\|_2 \leq \varepsilon \|g_i\|_2$. Using the submultiplicativity of matrix norms,
\begin{align}
    \|e_{\mathrm{tot}}\|_2^2 & = \|E\boldsymbol{\lambda}\|_2^2 \leq \|E\|_2^2 \|\boldsymbol{\lambda}\|_2^2 \leq \|E\|_F^2 \|\boldsymbol{\lambda}\|_2^2  \nonumber
\\ &= \left(\sum_{i=1}^m \|e_i\|_2^2\right) \|\boldsymbol{\lambda}\|_2^2 \leq \varepsilon^2 S \|\boldsymbol{\lambda}\|_2^2, \nonumber
\end{align}
where the Frobenius norm bound follows from the column-wise structure of $E$. Taking square roots yields
\begin{equation}\label{eq:residual_bound}
\|e_{\mathrm{tot}}\|_2 \leq \varepsilon \|\boldsymbol{\lambda}\|_2 \sqrt{S}.
\end{equation}

Next, we bound the subspace component. We have
\begin{equation}
    \|\tilde{g}_{\mathrm{tot}}\|_2^2 = \|\tilde{A}\boldsymbol{\lambda}\|_2^2 = \boldsymbol{\lambda}^\top \tilde{A}^\top \tilde{A} \boldsymbol{\lambda} \leq \|\boldsymbol{\lambda}\|_2^2 \cdot \lambda_{\max}(\tilde{A}^\top \tilde{A}). \nonumber
\end{equation}
Since $\tilde{A}^\top \tilde{A}$ and $\tilde{A}\tilde{A}^\top$ share the same nonzero eigenvalues, we have $\lambda_{\max}(\tilde{A}^\top \tilde{A}) = \lambda_{\max}(\tilde{A}\tilde{A}^\top)$. By assumption (S2),
\begin{align} 
\lambda_{\max}(\tilde{A}\tilde{A}^\top) & \leq (1 + \delta) \frac{\operatorname{trace}(\tilde{A}\tilde{A}^\top)}{r} \nonumber
\\ & = (1 + \delta) \frac{\sum_{i=1}^m \|\tilde{g}_i\|_2^2}{r} \leq (1 + \delta) \frac{S}{r}, \nonumber
\end{align}
where the last inequality uses $\|\tilde{g}_i\|_2 \leq \|g_i\|_2$. Therefore,
\begin{equation}\label{eq:subspace_bound}
\|\tilde{g}_{\mathrm{tot}}\|_2 \leq \|\boldsymbol{\lambda}\|_2 \sqrt{\frac{(1 + \delta)S}{r}}.
\end{equation}

Finally, substituting the bounds \eqref{eq:residual_bound} and \eqref{eq:subspace_bound} into the triangle inequality \eqref{eq:triangle},
\begin{align*}
\|g_{\mathrm{tot}}\|_2^2 
&\leq \left(\|\boldsymbol{\lambda}\|_2 \sqrt{\frac{(1 + \delta)S}{r}} + \varepsilon \|\boldsymbol{\lambda}\|_2 \sqrt{S}\right)^2 \\
&= \|\boldsymbol{\lambda}\|_2^2 S \left(\frac{1 + \delta}{r} + 2\varepsilon\sqrt{\frac{1 + \delta}{r}} + \varepsilon^2\right) \\
&= \|\boldsymbol{\lambda}\|_2^2 \cdot \frac{1 + \delta}{r} S \left(1 + 2\varepsilon\sqrt{\frac{r}{1 + \delta}} + \varepsilon^2 \frac{r}{1 + \delta}\right).
\end{align*}
Defining $\eta = 2\varepsilon\sqrt{\frac{r}{1 + \delta}} + \varepsilon^2 \frac{r}{1 + \delta} = O(\varepsilon\sqrt{r})$ when $\varepsilon$ is small, we obtain
$$\|g_{\mathrm{tot}}\|_2^2 \leq (1 + \eta) \cdot \frac{1 + \delta}{r} \|\boldsymbol{\lambda}\|_2^2 \sum_{i=1}^m \|g_i\|_2^2,$$
which completes the proof. $\hfill\square$
\end{Proof}

\textbf{Remark.} When task-specific gradients share a low-dimensional subspace with spectrally flat distribution, joint optimization achieves a $1/r$ reduction in gradient magnitude. In network tomography, forcing all tasks to operate through a shared latent representation $z$ naturally induces this low-rank structure, while diverse tomography objectives prevent gradient concentration, satisfying both \textit{\textbf{(S1)}} and \textit{\textbf{(S2)}}.

\section{Training Details}

\subsection*{A. Training Hyperparameters}

Table~\ref{tab:training_params} summarizes the hyperparameters in experiments.

\begin{table}[t]
\centering
\caption{Training Hyperparameters}
\label{tab:training_params}
\begin{tabular}{lc}
\hline
\textbf{Hyperparameter} & \textbf{Value} \\
\hline
Batch size & 64 \\
Number of epochs & 100 \\
Learning rate & $10^{-3}$ \\
Optimizer & Adam \\
$\lambda_{\text{1}}$ (alignment loss weight) & 1.0 \\
$\lambda_{\text{2}}$ (reconstruction loss weight) & 2.0 \\
$\lambda_{\text{3}}$ (Tasks loss weight) & 1.0 \\
Temperature $\tau$ (contrastive learning) & 0.7 \\
Latent dimension & 32 \\
Hidden dimensions & [128, 64] \\
Gradient clipping max norm & 1.0 \\
\hline
\end{tabular}
\end{table}

\subsection*{B. Model Architecture}

The PLATONT model consists of three parallel encoder-decoder pairs, one for each network indicator (delay, loss, bandwidth). Each encoder is a multi-layer perceptron that maps the input path-level measurements to a shared 32-dimensional latent representation. The encoder architecture follows a two-layer design with hidden dimensions of 128 and 64 units respectively. Each hidden layer is followed by ReLU activation, batch normalization, and dropout (0.1) for regularization. The decoder uses an adaptive aggregation mechanism that learns attention weights to combine latent representations from all three indicators before reconstruction. Specifically, each decoder contains an attention network that takes the concatenated latent representations $[z_{\text{delay}}, z_{\text{loss}}, z_{\text{bandwidth}}]$ as input and outputs three softmax-normalized weights. The final latent representation for decoding is computed as a weighted sum: $z_{\text{agg}} = w_1 z_{\text{delay}} + w_2 z_{\text{loss}} + w_3 z_{\text{bandwidth}}$. This adaptive mechanism allows each decoder to selectively leverage information from different indicator views.

\subsection*{C. Loss Normalization}

The reconstruction loss $\mathcal{L}_{\text{recon}} = \sum_{k \in \{\text{delay, loss, bw}\}} \mathcal{L}_k$ requires careful normalization to ensure balanced gradient contributions. We use Huber loss instead of MSE for robustness to outliers. The key challenge is that the three indicators have different natural scales and variances. Without normalization, delay MSE is around $10^1$, loss MSE is around $10^{-2}$, and bandwidth MSE is around $10^2$. To address this, we compute the standard deviation $\sigma_k$ for each indicator $k$ within each batch: $\sigma_k = \sqrt{\text{Var}(x_k^{\text{clean}})}$. Each reconstruction loss term is then normalized by its corresponding scale: $\mathcal{L}_k^{\text{normalized}} = \mathcal{L}_k / \max(\sigma_k, 10^{-6})$. This dynamic normalization ensures that all three loss terms contribute equally to the gradient during backpropagation, regardless of their absolute magnitudes. The normalized losses are then averaged to form the final reconstruction loss. We use $\lambda_{\text{recon}} = 2.0$ to emphasize reconstruction fidelity relative to alignment. Training uses AdamW optimizer with cosine annealing learning rate scheduling (warm restarts at epoch 10 with $T_{\text{mult}}=2$) and gradient clipping with max norm 1.0 for stability.

\begin{spacing}{0.8}  
\bibliographystyle{plain}
\renewcommand{\bibfont}{\footnotesize}
\bibliography{refs}

@article{vardi1996network_eng,
  title={Network tomography: Estimating source-destination traffic intensities from link data},
  author={Vardi, Yehuda},
  journal={Journal of the American statistical association},
  volume={91},
  number={433},
  pages={365--377},
  year={1996},
  publisher={Taylor \& Francis}
}

@inproceedings{dhamdhere2007netdiagnoser,
  title={NetDiagnoser: Troubleshooting network unreachabilities using end-to-end probes and routing data},
  author={Dhamdhere, Amogh and Teixeira, Renata and Dovrolis, Constantine and Diot, Christophe},
  booktitle={Proceedings of the 2007 ACM CoNEXT conference},
  pages={1--12},
  year={2007}
}

@article{chen2010network,
  title={Network tomography: Identifiability and fourier domain estimation},
  author={Chen, Aiyou and Cao, Jin and Bu, Tian},
  journal={IEEE Transactions on Signal Processing},
  volume={58},
  number={12},
  pages={6029--6039},
  year={2010},
  publisher={IEEE}
}

@inproceedings{chen2003network,
author = {Chen, Yan and Bindel, David and Katz, Randy H.},
title = {Tomography-based overlay network monitoring},
year = {2003},
isbn = {1581137737},
publisher = {Association for Computing Machinery},
address = {New York, NY, USA},
url = {https://doi.org/10.1145/948205.948233},
doi = {10.1145/948205.948233},
booktitle = {Proceedings of the 3rd ACM SIGCOMM Conference on Internet Measurement},
pages = {216–231},
numpages = {16},
location = {Miami Beach, FL, USA},
series = {IMC '03}
}

@article{Xi2006EstimatingNL,
  title={Estimating Network Loss Rates Using Active Tomography},
  author={Bowei Xi and George Michailidis and Vijayan N. Nair},
  journal={Journal of the American Statistical Association},
  year={2006},
  volume={101},
  pages={1430 - 1448},
}

@article{Duffield2006NetworkLT,
  title={Network loss tomography using striped unicast probes},
  author={Nick G. Duffield and Francesco Lo Presti and Vern Paxson and Donald F. Towsley},
  journal={IEEE/ACM Transactions on Networking},
  year={2006},
  volume={14},
  pages={697-710},
  url={https://api.semanticscholar.org/CorpusID:2790208}
}

@inproceedings{Nguyen2007NetworkLI,
  title={Network loss inference with second order statistics of end-to-end flows},
  author={H. Nguyen and Patrick Thiran},
  booktitle={ACM/SIGCOMM Internet Measurement Conference},
  year={2007},
}

@misc{Coates2000NetworkLI,
  title={Network loss inference using unicast end-to-end measurement},
  author={Coates, Mark and Nowak, Robert and others},
  booktitle={Proc. ITC Conf. IP Traffic, Modeling and Management},
  volume={28},
  year={2000}
}

@article{Singhal2007IdentifiabilityOF_od_1,
  title={Identifiability of flow distributions from link measurements with applications to computer networks},
  author={Harsh Singhal and George Michailidis},
  journal={Inverse Problems},
  year={2007},
  volume={23},
  pages={1821 - 1849},
  url={https://api.semanticscholar.org/CorpusID:15921093}
}

@article{Polverini2018RoutingPF_od_3,
  title={Routing Perturbation for Traffic Matrix Evaluation in a Segment Routing Network},
  author={Marco Polverini and Antonio Cianfrani and Marco Listanti and Andrea Baiocchi},
  journal={IEEE Transactions on Network and Service Management},
  year={2018},
  volume={15},
  pages={1645-1660},
  url={https://api.semanticscholar.org/CorpusID:56594366}
}

@article{mle2002,
  title={Maximum likelihood network topology identification from edge-based unicast measurements},
  author={Coates, Mark and others},
  journal={ACM SIGMETRICS Performance Evaluation Review},
  volume={30},
  number={1},
  pages={11--20},
  year={2002},
  publisher={ACM New York, NY, USA}
}

@inproceedings{EM2001-passive,
  title={Passive network tomography using EM algorithms},
  author={Tsang, Yolanda and Coates, Mark and Nowak, Robert},
  booktitle={2001 IEEE International Conference on Acoustics, Speech, and Signal Processing. Proceedings (Cat. No. 01CH37221)},
  volume={3},
  pages={1469--1472},
  year={2001},
  organization={IEEE}
}

@article{ni2009efficient,
  title={Efficient and dynamic routing topology inference from end-to-end measurements},
  author={Ni, Jian and others},
  journal={IEEE/ACM transactions on networking},
  volume={18},
  number={1},
  pages={123--135},
  year={2009},
  publisher={IEEE}
}

@misc{gnn-nt,
      title={Network Tomography with Path-Centric Graph Neural Network}, 
      author={Yuntong Hu and others},
      year={2025},
      eprint={2502.16430},
      archivePrefix={arXiv},
      primaryClass={cs.LG},
      url={https://arxiv.org/abs/2502.16430}, 
}

@article{Cceres1999MulticastbasedIO,
  title={Multicast-based inference of network-internal loss characteristics},
  author={Ram{\'o}n C{\'a}ceres and Nick G. Duffield and Joseph Horowitz and Donald F. Towsley},
  journal={IEEE Trans. Inf. Theory},
  year={1999},
  volume={45},
  pages={2462-2480},
  url={https://api.semanticscholar.org/CorpusID:840270}
}

@article{Tsang2003NetworkDT,
  title={Network delay tomography},
  author={Yolanda Tsang and Mark J. Coates and Robert D. Nowak},
  journal={IEEE Trans. Signal Process.},
  year={2003},
  volume={51},
  pages={2125-2136},
  url={https://api.semanticscholar.org/CorpusID:9735155}
}

@article{Bassi2022ImprovingDN_e1,
  title={Improving deep neural network generalization and robustness to background bias via layer-wise relevance propagation optimization},
  author={Pedro R.A.S. Bassi and Sergio S J Dertkigil and Andrea Cavalli},
  journal={Nature Communications},
  year={2022},
  volume={15},
  url={https://api.semanticscholar.org/CorpusID:251040577}
}

@article{Levin1998DYNAMICFE_e2,
  title={DYNAMIC FINITE ELEMENT MODEL UPDATING USING NEURAL NETWORKS},
  author={R. I. Levin and Nicholas A. J. Lieven},
  journal={Journal of Sound and Vibration},
  year={1998},
  volume={210},
  pages={593-607},
  url={https://api.semanticscholar.org/CorpusID:122740353}
}

@article{Duranthon2024AsymptoticGE_e3,
  title={Asymptotic generalization error of a single-layer graph convolutional network},
  author={O. Duranthon and Lenka Zdeborov'a},
  journal={ArXiv},
  year={2024},
  volume={abs/2402.03818},
  url={https://api.semanticscholar.org/CorpusID:267499817}
}

@article{PRH_2024,
  title={The Platonic Representation Hypothesis},
  author={Minyoung Huh and Brian Cheung and Tongzhou Wang and Phillip Isola},
  journal={ArXiv},
  year={2024},
  volume={abs/2405.07987},
  url={https://api.semanticscholar.org/CorpusID:269757765}
}

@article{Hjelm2018LearningDR,
  title={Learning deep representations by mutual information estimation and maximization},
  author={R. Devon Hjelm and Alex Fedorov and Samuel Lavoie-Marchildon and Karan Grewal and Adam Trischler and Yoshua Bengio},
  journal={ArXiv},
  year={2018},
  volume={abs/1808.06670},
  url={https://api.semanticscholar.org/CorpusID:52055130}
}

@inproceedings{ant-1-Caceres_Duffield_Moon_Towsley_2003,  
 title={Inference of internal loss rates in the MBone}, 
 url={http://dx.doi.org/10.1109/glocom.1999.832482}, 
 DOI={10.1109/glocom.1999.832482}, 
 booktitle={Seamless Interconnection for Universal Services. Global Telecommunications Conference. GLOBECOM’99. (Cat. No.99CH37042)}, 
 author={Caceres, R. and Duffield, N.G. and Moon, S.B. and Towsley, D.}, 
 year={2003}, 
 month={Jan}, 
 language={en-US} 
 }

@INPROCEEDINGS{ant-3-Sossalla,
  author={Sossalla, Peter and Rischke, Justus and Fitzek, Frank H. P.},
  booktitle={2022 International Conference on Information Networking (ICOIN)}, 
  title={Enhanced One-Way Delay Monitoring with OpenFlow}, 
  year={2022},
  volume={},
  number={},
  pages={171-176},
  doi={10.1109/ICOIN53446.2022.9687289}}

@article{ant-4-qiao, author = {Qiao, Yan and Jiao, Jun and Cui, Xinhong and Rao, Yuan}, title = {Robust Loss Inference in the Presence of Noisy Measurements and Hidden Fault Diagnosis}, year = {2020}, issue_date = {Feb. 2020}, publisher = {IEEE Press}, volume = {28}, number = {1}, issn = {1063-6692}, url = {https://doi.org/10.1109/TNET.2019.2948818}, doi = {10.1109/TNET.2019.2948818}, journal = {IEEE/ACM Trans. Netw.}, month = {feb}, pages = {43–56}, numpages = {14} }

@inproceedings{ant-5-Ghita_Nguyen_Kurant_Argyraki_Thiran_2010,  
 title={Netscope: Practical Network Loss Tomography}, 
 url={http://dx.doi.org/10.1109/infcom.2010.5461918}, 
 DOI={10.1109/infcom.2010.5461918}, 
 booktitle={2010 Proceedings IEEE INFOCOM}, 
 author={Ghita, Denisa and Nguyen, Hung and Kurant, Maciej and Argyraki, Katerina and Thiran, Patrick}, 
 year={2010}, 
 month={Mar}, 
 language={en-US} 
 }

@article{add-new-kolar2020distributed,
  title={Distributed network tomography applied to stochastic delay profile estimation},
  author={Kolar, Jakub and Sykora, Jan and Spagnolini, Umberto and others},
  journal={Radioengineering},
  volume={29},
  number={1},
  pages={189--196},
  year={2020}
}

@article{NBT_Duffield_2006,  
 title={Network Tomography of Binary Network Performance Characteristics}, 
 url={http://dx.doi.org/10.1109/tit.2006.885460}, 
 DOI={10.1109/tit.2006.885460}, 
 journal={IEEE Transactions on Information Theory}, 
 author={Duffield, Nick}, 
 year={2006}, 
 month={Dec}, 
 pages={5373–5388}, 
 language={en-US} 
 }

@inproceedings{nbt-1-Duffield_2003,  
 title={Simple network performance tomography}, 
 url={http://dx.doi.org/10.1145/948205.948232}, 
 DOI={10.1145/948205.948232}, 
 booktitle={Proceedings of the 2003 ACM SIGCOMM conference on Internet measurement  - IMC ’03}, 
 author={Duffield, Nick}, 
 year={2003}, 
 month={Jan}, 
 language={en-US} 
 }

@article{nbt-2-ogino,
author = {Ogino, Nagao and Kitahara, Takeshi and Arakawa, Shin’ichi and Hasegawa, Go and Murata, Masayuki},
year = {2018},
month = {04},
pages = {},
title = {Lightweight Boolean Network Tomography Based on Partition of Managed Networks},
volume = {26},
journal = {Journal of Network and Systems Management},
doi = {10.1007/s10922-017-9416-1}
}

@ARTICLE{nbt-4-bartolini,
  author={Bartolini, Novella and He, Ting and Arrigoni, Viviana and Massini, Annalisa and Trombetti, Federico and Khamfroush, Hana},
  journal={IEEE/ACM Transactions on Networking}, 
  title={On Fundamental Bounds on Failure Identifiability by Boolean Network Tomography}, 
  year={2020},
  volume={28},
  number={2},
  pages={588-601},
  doi={10.1109/TNET.2020.2969523}}

@inproceedings{range_nt,
  title={Range tomography: combining the practicality of boolean tomography with the resolution of analog tomography},
  author={Zarifzadeh, Sajjad and Gowdagere, Madhwaraj and Dovrolis, Constantine},
  booktitle={Proceedings of the 2012 Internet Measurement Conference},
  pages={385--398},
  year={2012}
}

@inproceedings{add-new-nt-dl-sartzetakis2022machine,
  title={Machine learning network tomography with partial topology knowledge and dynamic routing},
  author={Sartzetakis, Ippokratis and Varvarigos, Emmanouel},
  booktitle={GLOBECOM 2022-2022 IEEE Global Communications Conference},
  pages={4922--4927},
  year={2022},
  organization={IEEE}
}

@inproceedings{add-new-nt-dl-tao2023network,
  title={Network tomography and reinforcement learning for efficient routing},
  author={Tao, Xu and Silvestri, Simone},
  booktitle={2023 IEEE 20th International Conference on Mobile Ad Hoc and Smart Systems (MASS)},
  pages={384--389},
  year={2023},
  organization={IEEE}
}

@misc{add-new-nt-dl-hudeepnt,
      title={Network Tomography with Path-Centric Graph Neural Network}, 
      author={Yuntong Hu and Junxiang Wang and Liang Zhao},
      year={2025},
      eprint={2502.16430},
      archivePrefix={arXiv},
      primaryClass={cs.LG},
      url={https://arxiv.org/abs/2502.16430}, 
}

@misc{ma2020neuralnetworktomography,
      title={Neural Network Tomography}, 
      author={Liang Ma and Ziyao Zhang and Mudhakar Srivatsa},
      year={2020},
      eprint={2001.02942},
      archivePrefix={arXiv},
      primaryClass={cs.NI},
      url={https://arxiv.org/abs/2001.02942}, 
}

@inproceedings{chen2017efficient,
  title={Efficient hierarchical traffic measurement in software-defined datacenter networks},
  author={Chen, Shiping and Zhou, You and Chen, Shigang},
  booktitle={2017 IEEE 10th International Conference on Cloud Computing (CLOUD)},
  pages={163--170},
  year={2017},
  organization={IEEE}
}

@inproceedings{xu2019lightweight,
  title={Lightweight flow distribution for collaborative traffic measurement in software defined networks},
  author={Xu, Hongli and Chen, Shigang and Ma, Qianpiao and Huang, Liusheng},
  booktitle={IEEE INFOCOM 2019-IEEE Conference on Computer Communications},
  pages={1108--1116},
  year={2019},
  organization={IEEE}
}

@inproceedings{zhou2018highly,
  title={Highly compact virtual active counters for per-flow traffic measurement},
  author={Zhou, You and Zhou, Yian and Chen, Shigang and Zhang, Youlin},
  booktitle={IEEE INFOCOM 2018-IEEE Conference on Computer Communications},
  pages={1--9},
  year={2018},
  organization={IEEE}
}

@inproceedings{zhou2024correlation,
  title={Correlation Analysis for Exploring Large-Scale Latency Variability in WANs},
  author={Zhou, Ying and Wu, Mingzhen and Song, Fei},
  booktitle={2024 IEEE International Performance, Computing, and Communications Conference (IPCCC)},
  pages={1--6},
  year={2024},
  organization={IEEE}
}

@article{zhang2003fast,
  title={Fast accurate computation of large-scale IP traffic matrices from link loads},
  author={Zhang, Yin and Roughan, Matthew and Duffield, Nick and Greenberg, Albert},
  journal={ACM SIGMETRICS Performance Evaluation Review},
  volume={31},
  number={1},
  pages={206--217},
  year={2003},
  publisher={ACM New York, NY, USA}
}

@inproceedings{erramill2006independent,
  title={An independent-connection model for traffic matrices},
  author={Erramill, Vijayi and Crovella, Mark and Taft, Nina},
  booktitle={Proceedings of the 6th ACM SIGCOMM conference on Internet measurement},
  pages={251--256},
  year={2006}
}

@article{mardani2015estimating,
  title={Estimating traffic and anomaly maps via network tomography},
  author={Mardani, Morteza and Giannakis, Georgios B},
  journal={IEEE/ACM transactions on networking},
  volume={24},
  number={3},
  pages={1533--1547},
  year={2015},
  publisher={IEEE}
}

@inproceedings{soule2005traffic,
  title={Traffic matrices: balancing measurements, inference and modeling},
  author={Soule, Augustin and Lakhina, Anukool and Taft, Nina and Papagiannaki, Konstantina and Salamatian, Kave and Nucci, Antonio and Crovella, Mark and Diot, Christophe},
  booktitle={Proceedings of the 2005 ACM SIGMETRICS international conference on Measurement and modeling of computer systems},
  pages={362--373},
  year={2005}
}

@article{jiang2009garch,
  title={GARCH model-based large-scale IP traffic matrix estimation},
  author={Jiang, Dingde and Hu, Guangmin},
  journal={IEEE Communications Letters},
  volume={13},
  number={1},
  pages={52--54},
  year={2009},
  publisher={IEEE}
}

@article{xu2021learning,
  title={Learning based methods for traffic matrix estimation from link measurements},
  author={Xu, Shenghe and Kodialam, Murali and Lakshman, TV and Panwar, Shivendra S},
  journal={IEEE Open Journal of the Communications Society},
  volume={2},
  pages={488--499},
  year={2021},
  publisher={IEEE}
}

@article{liu2018tomogravity,
  title={Tomogravity space based traffic matrix estimation in data center networks},
  author={Liu, Guiyan and Guo, Songtao and Zhao, Quanjun and Yang, Yuanyuan},
  journal={Future Generation Computer Systems},
  volume={86},
  pages={39--50},
  year={2018},
  publisher={Elsevier}
}

@inproceedings{zhao2006robust,
  title={Robust traffic matrix estimation with imperfect information: Making use of multiple data sources},
  author={Zhao, Qi and Ge, Zihui and Wang, Jia and Xu, Jun},
  booktitle={Proceedings of the joint international conference on Measurement and modeling of computer systems},
  pages={133--144},
  year={2006}
}

@article{kumar2020multi,
  title={A multi-view subspace learning approach to internet traffic matrix estimation},
  author={Kumar, Awnish and Vidyapu, Sandeep and Saradhi, Vijaya V and Tamarapalli, Venkatesh},
  journal={IEEE Transactions on Network and Service Management},
  volume={17},
  number={2},
  pages={1282--1293},
  year={2020},
  publisher={IEEE}
}

@article{tian2018sdn,
  title={An SDN-based traffic matrix estimation framework},
  author={Tian, Yang and Chen, Weiwei and Lea, Chin-Tau},
  journal={IEEE Transactions on Network and Service Management},
  volume={15},
  number={4},
  pages={1435--1445},
  year={2018},
  publisher={IEEE}
}

@inproceedings{c-13,
  title={Optimal Positive Generation via Latent Transformation for Contrastive Learning},
  author={Yinqi Li and Hong Chang and Bingpeng Ma and S. Shan and Xilin Chen},
  booktitle={Neural Information Processing Systems},
  year={2022},
  url={https://api.semanticscholar.org/CorpusID:258509181}
}

@article{C-12,
  title={Towards a Unified Framework of Contrastive Learning for Disentangled Representations},
  author={Stefan Matthes and Zhiwei Han and Hao Shen},
  journal={ArXiv},
  year={2023},
  volume={abs/2311.04774},
  url={https://api.semanticscholar.org/CorpusID:265050622}
}

@article{c-14,
  title={Contrastive-Equivariant Self-Supervised Learning Improves Alignment with Primate Visual Area IT},
  author={Thomas E. Yerxa and Jenelle Feather and Eero P. Simoncelli and SueYeon Chung},
  journal={Advances in neural information processing systems},
  year={2024},
  volume={37},
  pages={
          96045-96070
        },
  url={https://api.semanticscholar.org/CorpusID:276260023}
}

@misc{c-15,
  title={Contrastive Learning for Correlating Network Incidents},
  author={Jeremias Dotterl},
  journal={arXiv preprint arXiv:2509.24446},
  year={2025}
}

@article{c-16,
  title={Self-Supervised Contrastive Learning for Videos using Differentiable Local Alignment},
  author={Keyne Oei and Amr Gomaa and Anna Maria Feit and Jo{\~a}o Marcelo Evangelista Belo},
  journal={ArXiv},
  year={2024},
  volume={abs/2409.04607},
  url={https://api.semanticscholar.org/CorpusID:272525318}
}

@inproceedings{c-21,
  title={Latent Multi-Task Architecture Learning},
  author={Sebastian Ruder and Joachim Bingel and Isabelle Augenstein and Anders S{\o}gaard},
  booktitle={AAAI Conference on Artificial Intelligence},
  year={2017},
  url={https://api.semanticscholar.org/CorpusID:115985550}
}

@article{c-23,
  title={You Only Learn One Representation: Unified Network for Multiple Tasks},
  author={Chien-Yao Wang and I-Hau Yeh and Hongpeng Liao},
  journal={J. Inf. Sci. Eng.},
  year={2021},
  volume={39},
  pages={691-709},
  url={https://api.semanticscholar.org/CorpusID:234334311}
}

@article{c-27,
  title={PixelBytes: Catching Unified Representation for Multimodal Generation},
  author={Fabien Furfaro},
  journal={ArXiv},
  year={2024},
  volume={abs/2410.01820},
  url={https://api.semanticscholar.org/CorpusID:273098354}
}

@misc{chen2020simpleframeworkcontrastivelearning,
      title={A Simple Framework for Contrastive Learning of Visual Representations}, 
      author={Ting Chen and Simon Kornblith and Mohammad Norouzi and Geoffrey Hinton},
      year={2020},
      eprint={2002.05709},
      archivePrefix={arXiv},
      primaryClass={cs.LG},
      url={https://arxiv.org/abs/2002.05709}, 
}

@misc{wu2018unsupervisedfeaturelearningnonparametric,
      title={Unsupervised Feature Learning via Non-Parametric Instance-level Discrimination}, 
      author={Zhirong Wu and Yuanjun Xiong and Stella Yu and Dahua Lin},
      year={2018},
      eprint={1805.01978},
      archivePrefix={arXiv},
      primaryClass={cs.CV},
      url={https://arxiv.org/abs/1805.01978}, 
}

@article{Malboubi2018OptimalCoherentNI,
  title={Optimal-Coherent Network Inference (OCNI): Principles and Applications},
  author={Mehdi Malboubi},
  journal={IEEE Transactions on Network and Service Management},
  year={2018},
  volume={15},
  pages={811-824},
  url={https://api.semanticscholar.org/CorpusID:49187255}
}

@article{du2024identificationpathcongestionstatus,
title = {Identification of path congestion status for network performance tomography using deep spatial–temporal learning},
journal = {Computer Communications},
volume = {239},
pages = {108194},
year = {2025},
issn = {0140-3664},
doi = {https://doi.org/10.1016/j.comcom.2025.108194},
url = {https://www.sciencedirect.com/science/article/pii/S0140366425001513},
author = {Chengze Du and others},
}

@InProceedings{secureNT,
author={Du, Chengze and Shi, Jibin and others},
title={SecureNT: Smart Topology Obfuscation for Privacy-Aware Network Monitoring},
booktitle={Advanced Intelligent Computing Technology and Applications},
year={2025},
publisher={Springer Nature Singapore},
doi={https://doi.org/10.1007/978-981-96-9914-8_9},
address={Singapore},
pages={101--112},
isbn={978-981-96-9914-8}
}

@article{du2025roto,
title={RoTO: Robust Topology Obfuscation Against Tomography Inference Attacks},
author={Du, Chengze and Xu, Heng and Yu, Zhiwei and Zhou, Ying and Meng, Zili and Li, Jialong},
journal={arXiv preprint arXiv:2508.12852},
year={2025},
url={https://arxiv.org/abs/2508.12852}, 
}

@misc{Donsker1975AsymptoticEO,
  title={Asymptotic evaluation of certain Markov process expectations for large time},
  author={Monroe D. Donsker and S. R. S. Varadhan},
  year={1975}
}

@article{He2019MomentumCF,
  title={Momentum Contrast for Unsupervised Visual Representation Learning},
  author={Kaiming He and Haoqi Fan and Yuxin Wu and Saining Xie and Ross B. Girshick},
  journal={2020 IEEE/CVF Conference on Computer Vision and Pattern Recognition (CVPR)},
  year={2019},
  pages={9726-9735},
  url={https://api.semanticscholar.org/CorpusID:207930212}
}

@misc{sergazinov2025ppdd,
      title={A spectral method for multi-view subspace learning using the product of projections}, 
      author={Renat Sergazinov and Armeen Taeb and Irina Gaynanova},
      year={2025},
      eprint={2410.19125},
      archivePrefix={arXiv},
      primaryClass={stat.ML},
      url={https://arxiv.org/abs/2410.19125}, 
}

@inproceedings{Varga2010OMNeT,
  title={OMNeT++},
  author={Andr{\'a}s Varga},
  booktitle={Modeling and Tools for Network Simulation},
  year={2010},
  url={https://api.semanticscholar.org/CorpusID:263881384}
}

@article{topohub,
title = {TopoHub: A repository of reference Gabriel graph and real-world topologies for networking research},
journal = {SoftwareX},
volume = {24},
pages = {101540},
year = {2023},
issn = {2352-7110},
doi = {https://doi.org/10.1016/j.softx.2023.101540},
url = {https://www.sciencedirect.com/science/article/pii/S2352711023002364},
author = {Piotr Jurkiewicz},
}

@inproceedings{
du2025semitoneaware,
title={Semitone-Aware Fourier Encoding: A Music-Structured Approach to Audio-Text Alignment},
author={Chengze Du and JinYang Zhang and Wenxin Zhang},
booktitle={AI for Music Workshop},
year={2025},
url={https://openreview.net/forum?id=KwKfcSGZhF}
}

@book{franklin2000matrix,
  title={Matrix Theory},
  author={Franklin, J.N.},
  isbn={9780486411798},
  lccn={99058316},
  series={Dover books on mathematics},
  url={https://books.google.com/books?id=-eNGCbdaZBIC},
  year={2000},
  publisher={Dover Publications}
}

@ARTICLE{11119637,
  author={Li, Jialong and De Marchi, Federico and Lei, Yiming and Joshi, Raj and Chandrasekaran, Balakrishnan and Xia, Yiting},
  journal={IEEE Transactions on Networking}, 
  title={Unlocking Diversity of Fast-Switched Optical Data Center Networks With Unified Routing}, 
  year={2025},
  volume={},
  number={},
  pages={1-16},
  doi={10.1109/TON.2025.3590083}}

@inproceedings{10.1145/3651890.3672245,
author = {Li, Jialong and Gong, Haotian and De Marchi, Federico and Gong, Aoyu and Lei, Yiming and Bai, Wei and Xia, Yiting},
title = {Uniform-Cost Multi-Path Routing for Reconfigurable Data Center Networks},
year = {2024},
isbn = {9798400706141},
publisher = {Association for Computing Machinery},
address = {New York, NY, USA},
url = {https://doi.org/10.1145/3651890.3672245},
doi = {10.1145/3651890.3672245},
booktitle = {Proceedings of the ACM SIGCOMM 2024 Conference},
pages = {433–448},
numpages = {16},
location = {Sydney, NSW, Australia},
series = {ACM SIGCOMM '24}
}

@inproceedings{Papagiannaki2004ADA,
  title={A distributed approach to measure IP traffic matrices},
  author={Konstantina Papagiannaki and Nina Taft and Anukool Lakhina},
  booktitle={ACM/SIGCOMM Internet Measurement Conference},
  year={2004},
  url={https://api.semanticscholar.org/CorpusID:15912803}
}

@INPROCEEDINGS{du2025vaes,
    author = {Du, Chengze and others},
    booktitle = {2025 International Joint Conference on Neural Networks (IJCNN)},
    title = {GuidedLatent: Defending VAEs against Membership Inference Attacks via Distribution-Guided Privacy},
    year = {2025},
    pages = {1-8},
    doi = {10.1109/IJCNN64981.2025.11227775}
}
\end{spacing}

\end{document}